\documentclass{article}

\PassOptionsToPackage{table}{xcolor}

\usepackage{microtype}
\usepackage{graphicx}
\usepackage{subcaption}
\usepackage{booktabs} 
\usepackage[most]{tcolorbox}
\tcbuselibrary{listings,skins,breakable,minted}
\usepackage{tabularx}
\usepackage[table]{xcolor}
\definecolor{inputbg}{gray}{0.93}
\newcommand{\hky}[1]{{\setlength\fboxsep{1pt}\colorbox{yellow!25}{\textbf{#1}}}}
\newcommand{\hkg}[1]{{\setlength\fboxsep{1pt}\colorbox{green!15}{\textbf{#1}}}}

\newcommand{\llk}{$\text{LL}_\textsc{Key}$}
\newcommand{\llv}{$\text{LL}_\textsc{Value}$}
\newcommand{\tl}{TL}
\newcommand{\zo}{ZO}
\newcommand{\tc}[1]{TC$_{a=#1}$}
\newcommand{\qlk}{$\text{QL}_\textsc{Key}$}
\newcommand{\qlv}{$\text{QL}_\textsc{Value}$}
\usepackage{multirow}

\usepackage{xurl}
\usepackage{hyperref}
\usepackage{enumitem}

\usepackage[preprint]{icml2026}

\usepackage{amsmath}
\usepackage{amssymb}
\usepackage{mathtools}
\usepackage{amsthm}

\usepackage[capitalize,noabbrev]{cleveref}

\theoremstyle{plain}

\theoremstyle{definition}

\theoremstyle{remark}

\usepackage[textsize=tiny]{todonotes}

\icmltitlerunning{Query Lens: Interpreting Sparse Key-Value Features with Indirect Effects}

\begin{document}

\twocolumn[
  \icmltitle{Query Lens: Interpreting Sparse Key-Value Features with Indirect Effects}



  \icmlsetsymbol{equal}{*}

  \begin{icmlauthorlist}
    \icmlauthor{Hwiyeong Lee}{sch}
    \icmlauthor{Ingyu Bang}{sch}
    \icmlauthor{Uiji Hwang}{sch}
    \icmlauthor{Hyelim Lim}{sch}
    \icmlauthor{Taeuk Kim}{sch}
  \end{icmlauthorlist}

  \icmlaffiliation{sch}{Hanyang University, Seoul, Republic of Korea}

  \icmlcorrespondingauthor{Taeuk Kim}{kimtaeuk@hanyang.ac.kr}

  \icmlkeywords{Machine Learning, ICML}

  \vskip 0.3in
]



\printAffiliationsAndNotice{}  

\begin{abstract}
While sparse autoencoders provide features more interpretable than individual neurons, reliably characterizing them remains challenging.
We propose Query Lens, which extends Logit Lens to enable more comprehensive and faithful interpretations of sparse features. 
By jointly considering encoder-side key features and decoder-side value features, we identify both the inputs that activate a feature and the outputs it promotes.
We also account for indirect, module-mediated effects that arise when the feature is processed by downstream modules, going beyond the direct effect captured by Logit Lens.
In experiments, we find that Query Lens yields coherent token signatures for features that remain uninterpretable under Logit Lens.
Finally, we propose the Subspace Channel Hypothesis, suggesting that downstream modules read features through layer-specific subspaces.
\end{abstract}

\section{Introduction}

Explaining the inner workings of large language models (LLMs) remains a central challenge in the field of mechanistic interpretability. 
A core objective of this line of work \cite{bau2018identifying, mu2021compositionalexplanationsneurons,dai-etal-2022-knowledge,park2025geometrycategoricalhierarchicalconcepts} is to assign human-interpretable descriptions to the LLMs' internal representations, a.k.a. features. 
Recent progress in sparse dictionary learning, particularly through sparse autoencoders (SAEs), has accelerated this research direction by providing a more tractable target of analysis: internal activations represented as sparse combinations of dictionary elements \cite{cunningham2023sparseautoencodershighlyinterpretable}.

\begin{figure}[t]
  \centering
  \begin{subfigure}[t]{0.95\linewidth}
    \centering
    \includegraphics[width=\linewidth]{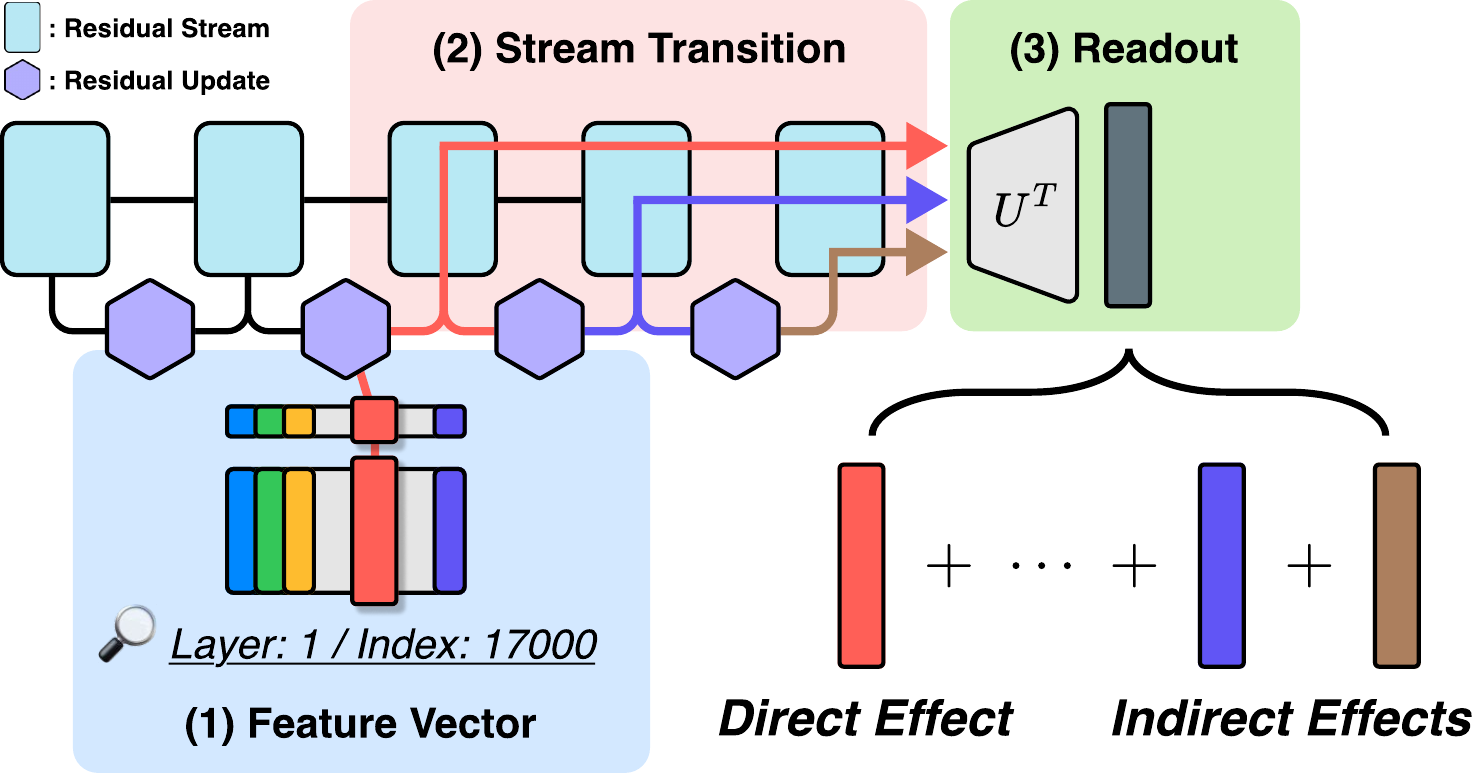}
    \caption{Schematic view of Residual Stream Dynamics.}
    \label{fig:main1}
  \end{subfigure}

  \vspace{0.6em}
  \begin{subfigure}[t]{0.95\linewidth}
    \centering
    \includegraphics[width=\linewidth]{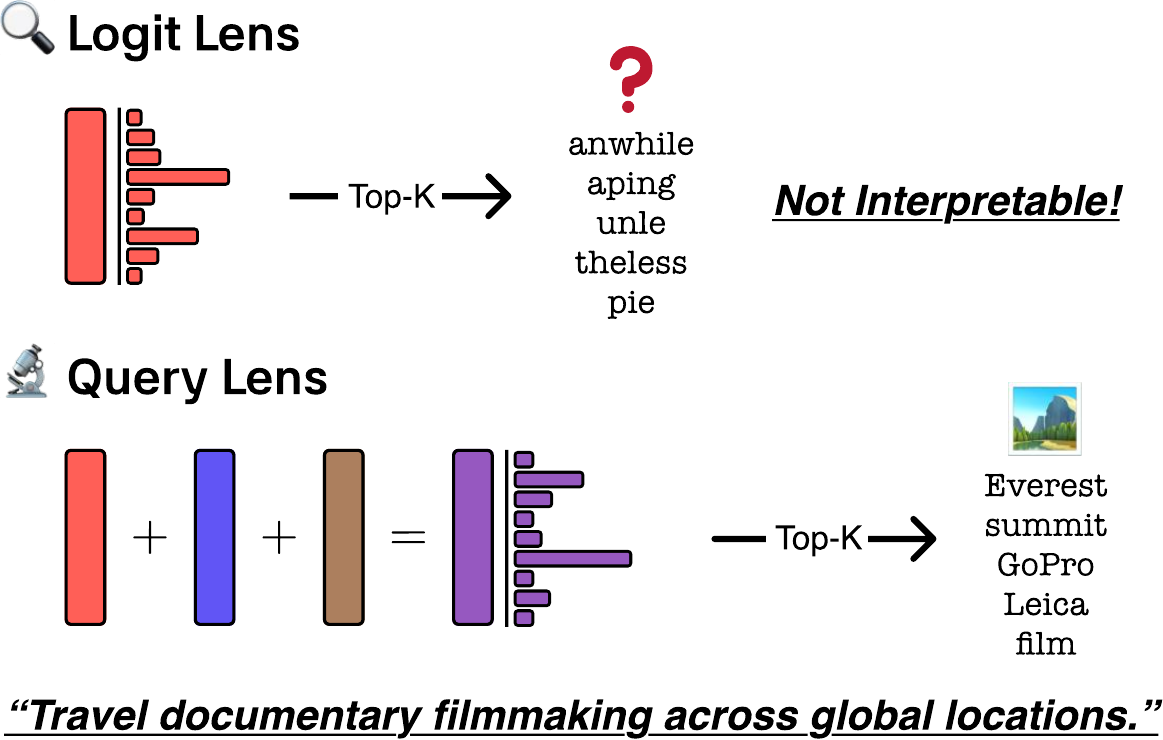}
    \caption{Comparison of Logit Lens and Query Lens.}
    \label{fig:main2}
  \end{subfigure}
  \caption{Overview of \textbf{Query Lens}. \textbf{(a)} A feature written into the residual stream is read as a \emph{query} by downstream modules, producing \emph{indirect effects}. \textbf{(b)} Logit Lens projects features directly into vocabulary space and misses these indirect effects. Query Lens accounts for them and provides a more faithful interpretation.}
  \label{fig:main}
\end{figure}

A common approach to characterizing SAE features is data-driven: the model is run on large corpora to identify inputs that strongly activate a target feature, and the resulting high-activation contexts are used to infer the feature's semantics.
Although a large body of prior work \cite{bills2023language,bricken2023monosemanticity,choi2024automatic,paulo2025automaticallyinterpretingmillionsfeatures} relies on this practice, it suffers from two notable limitations.
First, securing feature-sensitive examples typically requires exhaustive model runs over large corpora; in some cases, access to the underlying data is even infeasible due to privacy constraints \cite{dar-etal-2023-analyzing}.
Second, input-conditioned descriptions alone are not output-grounded: they often fail to sufficiently capture a feature's causal effect on the model's generation, limiting their reliability for downstream applications such as steering \cite{gur-arieh-etal-2025-enhancing}.

A prominent alternative for interpreting SAE features is to project them directly into the vocabulary space, primarily by applying Logit Lens \cite{nostalgebraist2020interpreting,bloom2024understandingfeatureslogitlens}.
While this method avoids the complexity of collecting activating contexts and provides an output-grounded summary, it nonetheless comes with two fundamental drawbacks.
(1) \textbf{Completeness:} While Logit Lens  captures how a feature direction affect the output logits, it does not explain which inputs originally activate the feature, i.e., the input-side causality.
(2) \textbf{Faithfulness:} A large portion of SAE features remain elusive under Logit Lens. 
In particular, many features, especially in earlier layers, exhibit diffuse token patterns or are dominated by uninterpretable tokens, rather than converging to a coherent semantic concept.

In this work, we aim to address the two aforementioned issues. 
To improve completeness, we extend probing beyond SAE decoders to include SAE encoders that have received comparatively less attention.
Adopting the key-value memory view \cite{geva-etal-2021-transformer}, we refer to encoder features as \textit{key features} and decoder features as \textit{value features}. 
Concretely, key features compute activations in response to the input, whereas value features are added to the residual stream weighted by these activations. 
Accordingly, this structure maps directly onto two aspects of a feature's causal role: key features characterize which inputs activate a feature, while value features specify which outputs it promotes.

In terms of faithfulness, we argue that by design Logit Lens reveals only a subset of viable input–feature and feature–output interactions. Specifically, when a feature direction is added to the residual stream, its effect on the output distribution decomposes into two components: a direct effect and an indirect effect.\footnote{While the discussion here centers on feature–output interactions, the same direct/indirect decomposition applies to input–feature effects.} 
The \textbf{direct effect} propagates through the residual stream to the output logits, whereas the \textbf{indirect effect} arises when downstream modules consume the feature in the residual stream, inducing additional changes in the output logits.
Logit Lens focuses on the direct effect while largely ignoring the indirect effect, leading to the prevalence of uninterpretable token distributions.

To this end, we propose \textbf{Query Lens}, a framework that extends Logit Lens to interpret SAE features in embedding space in a more comprehensive and faithful manner.
First, to characterize a feature's causal role on both the input and output sides, Query Lens adaptively switches between key and value features: the former captures what activates a feature, while the latter reveals what it promotes.
Furthermore, Query Lens accounts for both the direct and indirect effects of features, providing more reliable explanations of their functions across a wider range of cases.
In sum, these improvements enable token-level interpretations that more accurately reflect the causal footprint of SAE features.

By explicitly considering indirect effects, Query Lens further supports a mechanistic analysis of how features are converted and read by downstream modules in LLMs.
We observe that a feature produces markedly different impacts across Transformer components, despite being added to the residual stream as a static vector.
Motivated by this finding, we introduce the \textbf{Subspace Channel Hypothesis}.
We assume that the heterogeneous processing of a feature vector stems from selective readout: rather than uniformly consuming the full feature, each module extracts information from a low-dimensional subspace, termed a channel.
We investigate this phenomenon by learning low-rank linear maps from features to module responses, revealing that feature readout is mediated by layer-specific channels. Our code is available at \url{https://github.com/HYU-NLP/query-lens}.

\section{Background}
\subsection{Residual Stream View of Transformers}
We adopt the residual stream view of Transformer language models, where hidden states form a single stream updated only through residual additions \cite{elhage2021mathematical}. 
Each layer consists of two residual blocks, a multi-head self-attention block $R_{\text{A}}(\cdot)$ and an MLP block $R_{\text{M}}(\cdot)$, both of which add residual updates to the stream. 
Formally,
\[
h_{\text{mid}}^{l} = h_{\text{pre}}^{l} + R_{\text{A}}^{l}(h_{\text{pre}}^{l}), \quad
h_{\text{post}}^{l} = h_{\text{mid}}^{l} + R_{\text{M}}^{l}(h_{\text{mid}}^{l}).
\]
The residual stream then propagates to the next layer by setting $h_{\text{pre}}^{l+1} = h_{\text{post}}^{l}$.
We omit LayerNorm for brevity.

\subsection{Sparse Autoencoders}
A bottleneck in interpreting neural networks is that many neurons are \emph{polysemantic}, responding to multiple, unrelated explanations due to \emph{superposition} \cite{elhage2022superposition,bricken2023monosemanticity}. 
To disentangle such features into a more localized basis, recent work has adopted \textbf{sparse autoencoders (SAEs)}.
SAEs reconstruct the original residual stream vectors while encouraging sparse feature activations. 
Given a target vector $h_{\text{post}} \in \mathbb{R}^{d_m}$, an SAE computes
\begin{equation}
\label{eq:sae-computation}
\hat{h}_{\text{post}}
= W_{\text{dec}} f\!\left(W_{\text{enc}}^\top h_{\text{post}}\right),
\end{equation}
where $W_{\text{enc}}, W_{\text{dec}} \in \mathbb{R}^{d_m \times d_{\text{dict}}}$ denote the encoder and decoder weight matrices, $f(\cdot)$ is a pointwise nonlinearity, and $\hat{h}_{\text{post}}$ denotes the SAE reconstruction of $h_{\text{post}}$. 
By using an overcomplete dictionary ($d_{\text{dict}} \gg d_m$) with sparse activations, SAEs learn features that are more monosemantic and thus more characterizable than individual neurons, which motivates their analysis.

\subsection{Key-Value Memory}
We employ the \emph{key–value memory} view of the MLP block \cite{geva-etal-2021-transformer} and extend it to sparse dictionaries. 
The SAE computation (Eq.~\eqref{eq:sae-computation}) can be written as a sum of \emph{sub-updates}. 
Let $k_i$ and $v_i$ denote the $i$-th columns of $W_{\text{enc}}$ and $W_{\text{dec}}$. Then, the SAE reconstruction can be decomposed as
\begin{equation}
\hat{h}_{\text{post}}
= \sum_{i=1}^{d_{\text{dict}}} a_i\!\left(h_{\text{post}}\right)\, v_i,
\
a_i\!\left(h_{\text{post}}\right)
= f\!\left(\langle h_{\text{post}}, k_i \rangle\right).
\label{eq:sae-decomp}
\end{equation}
Each sub-update is computed by taking an inner product between the input vector $h_{\text{post}}$ and a column vector from the encoder ($k_i$) to obtain a scalar activation ($a_i$), and using it to weight the corresponding vector from the decoder ($v_i$).

This yields an attention-style analogy: encoder column vectors $\{k_i\}$ serve as \emph{key features} that produce sparse activations from the input, while decoder column vectors $\{v_i\}$ serve as \emph{value features} combined using these activations.

\section{Query Lens}

\subsection{Key Concepts and Limitations of Logit Lens}

Logit Lens \cite{nostalgebraist2020interpreting} was originally proposed as a method for inspecting \emph{intermediate hidden states} in a Transformer by asking what the model would predict if decoding were performed from an intermediate layer.
It projects a residual stream vector into vocabulary space as
\begin{equation*}
y^{l} = U^{\top} h_{\text{post}}^{l} \in \mathbb{R}^{|V|},
\label{eq:logit-lens}
\end{equation*}
where $U \in \mathbb{R}^{d_m \times |V|}$ is the unembedding matrix, and $|V|$ is the size of the vocabulary $V$.

Prior work has further applied Logit Lens to model \emph{parameters} \cite{geva-etal-2021-transformer,geva-etal-2022-transformer}.
Since $h_{\text{post}}^{l}$ can be estimated using a sum of sub-updates as in Eq.~\eqref{eq:sae-decomp}, projecting a single sub-update isolates its contribution in vocabulary space:
\[
U^{\top}\!\left(a_i^{l}(h_{\text{post}}^{l})\, v_i^{l}\right)
= a_i^{l}(h_{\text{post}}^{l})\, U^{\top} v_i^{l}.
\]
Because the scalar activation $a_i^{l}(h_{\text{post}}^{l})$ only rescales logits and does not change their ranking, the static value vector determines the promoted token signature via $U^{\top} v_i^{l}$.

Despite its widespread use, interpreting SAE features with Logit Lens suffers from two key limitations. 
First, many SAE features yield uninterpretable tokens under Logit Lens. 
We attribute this in part to what Logit Lens measures: it primarily reflects the \textbf{direct effect} of adding a feature direction to the residual stream, while ignoring \textbf{indirect effects} that arise when downstream modules consume the perturbed stream and propagate its influence through their computations.
Second, current practice places little emphasis on key features in SAE encoders. 
Although a few seminal studies \cite{geva-etal-2021-transformer,dar-etal-2023-analyzing} analyze projections of MLP key vectors, this direction of work remains weakly connected to the dominant data-driven approach to SAE feature interpretation based on activating examples.

These limitations motivate the questions we address: how can we interpret a feature's causal role by (1) characterizing what activates it and what it promotes, and (2) accounting for both direct effects and indirect, module-mediated effects?

\subsection{Residual Stream Dynamics}

We begin by framing this question using the residual stream.
Let $x$ denote the one-hot indicator of the input token, so that the input embedding is obtained via lookup as $e = Ex$, where $E \in \mathbb{R}^{d_{\text{m}} \times |V|}$ denotes the embedding matrix.
Let $a$ be a feature activation from the forward pass and $y$ the output logits, where $y = U^{\top} h$ and $U \in \mathbb{R}^{d_{\text{m}} \times |V|}$ is the unembedding matrix.
The activation depends on the input and the output depends on the activation, i.e., $x \mapsto a$ and $a \mapsto y$.
Our goal is to characterize how these dependencies are expressed along the residual stream by relating variations in the input $x$ to changes in the feature activation $a$, and variations in $a$ to changes in the output logits $y$.

\subsubsection{Forward Dynamics}

We first formalize how a local perturbation to a feature activation propagates forward to the output logits. 
Consider feature $i$ at layer $l \in \{1,\ldots,L\}$, with post-activation $a_{i}^{\,l}$, i.e., the scalar obtained after applying the nonlinearity. 
The change in the output logits $y$ induced by perturbing this activation from $a^{\ast}$ by $\Delta a$ can be approximated as
\begin{equation*}
y\!\left(a^{\ast}+\Delta a\right)
\approx
y\!\left(a^{\ast}\right)
+
\left.\frac{\partial y}{\partial a_{i}^{\,l}}\right|_{a_{i}^{\,l}=a^{\ast}}\Delta a.
\label{eq:forward-linearization-logit}
\end{equation*}
This is a first-order linearization, where the induced logit change per unit change in the scalar activation is fixed by $\partial y/\partial a_{i}^{\,l}$. From $y = U^{\top} h_{\text{post}}^{L}$ and $h_{\text{post}}^{l} \approx \sum_i a_{i}^{\,l}\, v_i^{\,l}$, and with chain rule, it can be expanded as
\begin{equation}
\frac{\partial y}{\partial a_{i}^{\,l}}
=
\frac{\partial y}{\partial h_{\text{post}}^{L}}
\,
\frac{\partial h_{\text{post}}^{L}}{\partial h_{\text{post}}^{l}}
\,
\frac{\partial h_{\text{post}}^{l}}{\partial a_{i}^{\,l}} = U^\top \frac{\partial h_{\text{post}}^{L}}{\partial h_{\text{post}}^{l}} v_i^l.
\label{eq:chain-rule-logit}
\end{equation}
The intermediate mapping from $h_{\text{post}}^{l}$ to $h_{\text{post}}^{L}$ can be written as a product of Jacobians across downstream residual blocks.
To be specific, for each layer $k$, differentiating the equation from the definition of residual update gives
\begin{equation}
\frac{\partial h_{\text{mid}}^{k}}{\partial h_{\text{pre}}^{k}}
=
I + J_{\text{A}}^{k},
\qquad
\frac{\partial h_{\text{post}}^{k}}{\partial h_{\text{mid}}^{k}}
=
I + J_{\text{M}}^{k},
\label{eq:stream-transition-jacobians}
\end{equation}
where $J_{\text{A}}^{k}\coloneqq \partial R_{\text{A}}^{k}/\partial h_{\text{pre}}^{k}$ and
$J_{\text{M}}^{k}\coloneqq \partial R_{\text{M}}^{k}/\partial h_{\text{mid}}^{k}$.\footnote{Each Jacobian is evaluated at the corresponding pre-module residual from the forward pass on a reference input $x$ (e.g., $J_{\text{M}}^{k}$ at $h_{\text{mid}}^{k}(x)$); we leave this notation implicit for brevity.}
Since $h_{\text{pre}}^{k} = h_{\text{post}}^{k-1}$, repeated use of the chain rule yields
\begin{equation}
\frac{\partial h_{\text{post}}^{L}}{\partial h_{\text{post}}^{l}}
=
\prod_{k=l+1}^{L}
\frac{\partial h_{\text{post}}^{k}}{\partial h_{\text{mid}}^{k}}
\frac{\partial h_{\text{mid}}^{k}}{\partial h_{\text{pre}}^{k}}
=
\prod_{k=l+1}^{L}
\bigl(I + J_{\text{M}}^{k}\bigr)\bigl(I + J_{\text{A}}^{k}\bigr).
\label{eq:post-to-post-product}
\end{equation}

Combining Eq.~\eqref{eq:chain-rule-logit} and Eq.~\eqref{eq:post-to-post-product} gives
\begin{equation*}
\frac{\partial y}{\partial a_{i}^{\,l}}
=
U^{\top}
\left[
\prod_{k=l+1}^{L}
\bigl(I + J_{\text{M}}^{k}\bigr)\bigl(I + J_{\text{A}}^{k}\bigr)
\right]
v_{i}^{\,l}.
\end{equation*}

We refer to $\partial y/\partial a_{i}^{\,l}$ as the \textbf{forward dynamics} of feature $(l,i)$: it quantifies the logit change induced by a unit change in the post-activation $a_{i}^{\,l}$.

\subsubsection{Backward Dynamics}

Similarly, we formalize how a feature activation responds to input perturbations.
Here, $a_i^{\,l}$ denotes the \emph{pre-activation}, i.e., the scalar before applying the nonlinearity, which we use as a proxy for the post-activation, since common activation functions used in SAEs can be non-differentiable; we justify this proxy in Appendix~\ref{app:pre-act-proxy}.

Consider perturbing the input token indicator $x$ from a reference value $x^{\ast}$ by $\Delta x$.
The induced change in the activation $a_i^{\,l}$ can be approximated as
\begin{equation*}
a_i^{\,l}\!\left(x^{\ast}+\Delta x\right)
\approx
a_i^{\,l}\!\left(x^{\ast}\right)
+
\left.\frac{\partial a_i^{\,l}}{\partial x}\right|_{x=x^{\ast}}
\Delta x.
\label{eq:backward-linearization-x}
\end{equation*}

From $h_{\text{pre}}^{1}=E x$ and $a_i^{\,l}=\langle h_{\text{post}}^{l}, k_i^{\,l}\rangle$, the chain rule gives,
\begin{equation}
\frac{\partial a_i^{\,l}}{\partial x}
=
\frac{\partial a_i^{\,l}}{\partial h_{\text{post}}^{l}}
\,
\frac{\partial h_{\text{post}}^{l}}{\partial h_{\text{pre}}^{1}}
\,
\frac{\partial h_{\text{pre}}^{1}}{\partial x} = (k_i^l)^\top 
\frac{\partial h_{\text{post}}^{l}}{\partial h_{\text{pre}}^{1}} E.
\label{eq:chain-rule-x}
\end{equation}
The remaining term $\partial h_{\text{post}}^{l}/\partial h_{\text{pre}}^{1}$ expands into a product of Jacobians over the prefix of residual blocks from $k=1$ to $l$. Concretely, using Eq.~\eqref{eq:stream-transition-jacobians} and noting that $h_{\text{pre}}^{k+1}=h_{\text{post}}^{k}$, repeated application of the chain rule gives
\begin{equation}
\frac{\partial h_{\text{post}}^{l}}{\partial h_{\text{pre}}^{1}}
=
\prod_{k=1}^{l}
\frac{\partial h_{\text{post}}^{k}}{\partial h_{\text{mid}}^{k}}
\frac{\partial h_{\text{mid}}^{k}}{\partial h_{\text{pre}}^{k}}
=
\prod_{k=1}^{l}
\bigl(I + J_{\text{M}}^{k}\bigr)\bigl(I + J_{\text{A}}^{k}\bigr),
\label{eq:pre1-to-postl-product}
\end{equation}
where $J_{\text{A}}^{k}$ and $J_{\text{M}}^{k}$ are defined as in Eq.~\eqref{eq:stream-transition-jacobians}. 
Combining Eq.~\eqref{eq:chain-rule-x} and Eq.~\eqref{eq:pre1-to-postl-product} yields
\begin{equation*}
\frac{\partial a_i^{\,l}}{\partial x}
=
(k_i^{\,l})^{\top}
\left[
\prod_{k=1}^{l}
\bigl(I + J_{\text{M}}^{k}\bigr)\bigl(I + J_{\text{A}}^{k}\bigr)
\right]
E.
\label{eq:da-dx-final}
\end{equation*}

We refer to $\partial a_i^{\,l}/\partial x$ as the \textbf{backward dynamics} of feature $(l,i)$: it specifies the direction in input space that most increases the feature's pre-activation.

\subsection{Understanding the Dynamics} 

The above derivations reveal an analogy between the forward and backward dynamics, which can be factored into three elements: (1) \textbf{Feature Vector}, (2) \textbf{Stream Transition}, and (3) \textbf{Readout}.
They describe what is injected into the residual stream, how the signal is transported across layers, and how it is expressed in vocabulary space, respectively.

\paragraph{Feature Vector.}
The feature vectors are recovered by unpacking the \emph{local} coupling between a feature activation and the residual stream at the same layer. 
Asking what the dictionary \emph{writes} to the stream and what it \emph{reads} from the stream amounts to identifying two local derivatives, which reduce directly to the dictionary vectors themselves: $\partial h_{\text{post}}^{l}/\partial a_i^{\,l} = v_i^{\,l}$ and $\partial a_i^{\,l}/\partial h_{\text{post}}^{l} = (k_i^{\,l})^{\top}$.
Thus, feature vectors are not auxiliary artifacts; they \emph{are} the dictionary's local read/write directions: value features govern \emph{outgoing} influence on the residual stream, while key features capture \emph{incoming} sensitivity to residual stream variations.

\paragraph{Stream Transition.}
The \emph{stream transition} denotes the Jacobians $\partial h_{\text{post}}^{l} / \partial h_{\text{pre}}^{1}$ and $\partial h_{\text{post}}^{L} / \partial h_{\text{post}}^{l}$, since they specify how a local read/write at layer $l$ is transported through the residual stream and expressed at its endpoints, $h_{\text{pre}}^{1}$ and $h_{\text{post}}^{L}$.
Specifically, this is a $d_{\mathrm{m}}\times d_{\mathrm{m}}$ mapping, and as Eq.~\eqref{eq:post-to-post-product} and Eq.~\eqref{eq:pre1-to-postl-product} show, it is expressed as a product of identity-plus-Jacobian factors: 
$\prod_k (I+J_{\mathrm{M}}^{k})(I+J_{\mathrm{A}}^{k})$ (abbreviated as $\prod_k (I+J^{k})$ hereafter).
Each term in the expansion of this product can be interpreted as a distinct computational path by which the local perturbation reaches an endpoint stream. 
For instance, the term $J_{\text{M}}^{5}$ corresponds to the path where the induced residual change is consumed by the layer-5 MLP block, modifying its output and in turn altering the residual stream that continues forward.

From this perspective, Logit Lens retains only the identity term in the stream transition, considering only the computation from a local perturbation to the endpoint stream, i.e., the \emph{direct effect}.
By including the remaining terms---which capture how downstream modules consume and reshape the perturbation, i.e., the \emph{indirect effects}---we obtain a more faithful transport of the local effect to the endpoint stream.

\paragraph{Readout.}
The readout defines how a local effect is expressed in vocabulary space, namely as a map $\mathbb{R}^{d_\text{m}}\to\mathbb{R}^{|V|}$.
Concretely, the endpoint readouts correspond to the embedding and unembedding matrices: $\partial h_{\text{pre}}^{1}/\partial x = E$ at the input endpoint and $\partial y/\partial h_{\text{post}}^{L} = U^{\top}$ at the output endpoint.

For the backward readout in particular, we modify $E$: our goal is not to interpret the transported direction at $h_{\text{pre}}^{1}$ itself, but to identify which specific input-token change from the current token would realize it. This motivates adjusting the readout in terms of \emph{token substitutions}.
Specifically, if $e_x$ induces a direction $\Delta h_{\text{pre}}^{1}$ at the input endpoint, we choose the substitute token whose embedding shift $(e_{x'}-e_x)$ best aligns with $\Delta h_{\text{pre}}^{1}$.
To implement this, we form a centered embedding matrix that represents substitution effects, $\widetilde{E}=E-e_x\mathbf{1}^{\top}$, where each column $\tilde e_t=e_t-e_x$ corresponds to the embedding change resulting from substituting the input token with $t$.
We then normalize each centered vector to unit norm, yielding a centered-normalized matrix $\widehat{E}$ with columns $\hat e_t=\tilde e_t/\lVert \tilde e_t\rVert_2$.
Reading out with $\widehat{E}$ therefore captures how well substituting each candidate token $t$ for the input $x$ matches the transported direction $\Delta h_{\text{pre}}^{1}$.\footnote{Because $x$ is discrete, no substitution is guaranteed to match $\Delta h_{\text{pre}}^{1}$; however, the features we interpret are empirically confirmed to activate on some tokens, so this is not a practical concern.}

\begin{table*}[t]
  \caption{Input and Output scores (\%) across methods on four model/SAE configurations. Values are the mean score over layers along with their respective 95\% confidence intervals. For each (configuration, score-type) the largest value is in \textbf{bold}.}
  \label{tab:main}
  \begin{center}
    \begin{footnotesize}
      \begin{tabular}{l
        >{\columncolor{inputbg}}r r
        >{\columncolor{inputbg}}r r
        >{\columncolor{inputbg}}r r
        >{\columncolor{inputbg}}r r}
        \toprule
        & \multicolumn{2}{c}{\textbf{GPT-2 Small}} & \multicolumn{2}{c}{\textbf{Gemma-3-270M}} & \multicolumn{2}{c}{\textbf{Gemma-3-1B}} & \multicolumn{2}{c}{\textbf{Qwen-3-1.7B}} \\
        \cmidrule(lr){2-3} \cmidrule(lr){4-5} \cmidrule(lr){6-7} \cmidrule(lr){8-9}
        & Input (\%) & Output (\%) & Input (\%) & Output (\%) & Input (\%) & Output (\%) & Input (\%) & Output (\%) \\
        \midrule
        \llk      & $\text{7.84}_{\pm \text{1.05}}$ & $\text{4.32}_{\pm \text{0.43}}$ & $\text{0.71}_{\pm \text{0.58}}$ & $\text{1.79}_{\pm \text{0.67}}$ & $\text{1.74}_{\pm \text{1.01}}$ & $\text{3.25}_{\pm \text{0.88}}$ & $\text{1.91}_{\pm \text{0.90}}$ & $\text{4.43}_{\pm \text{0.81}}$ \\
        \llv      & $\text{11.74}_{\pm \text{1.41}}$ & $\text{12.57}_{\pm \text{0.60}}$ & $\text{2.14}_{\pm \text{1.25}}$ & $\text{6.13}_{\pm \text{0.98}}$ & $\text{1.03}_{\pm \text{0.67}}$ & $\text{7.84}_{\pm \text{0.93}}$ & $\text{3.56}_{\pm \text{1.28}}$ & $\text{8.31}_{\pm \text{0.97}}$ \\
        \tl       & $\text{8.24}_{\pm \text{1.22}}$ & $\text{12.03}_{\pm \text{0.60}}$ & $\text{2.23}_{\pm \text{1.23}}$ & $\text{5.38}_{\pm \text{0.94}}$ & $\text{1.01}_{\pm \text{0.67}}$ & $\text{7.81}_{\pm \text{0.92}}$ & $\text{3.61}_{\pm \text{1.29}}$ & $\text{8.14}_{\pm \text{0.97}}$ \\
        \zo       & $\text{6.63}_{\pm \text{1.15}}$ & $\text{5.69}_{\pm \text{0.56}}$ & $\text{5.78}_{\pm \text{2.12}}$ & $\text{2.75}_{\pm \text{0.86}}$ & $\text{4.24}_{\pm \text{1.78}}$ & $\text{3.13}_{\pm \text{0.87}}$ & $\text{12.55}_{\pm \text{2.84}}$ & $\text{6.21}_{\pm \text{0.94}}$ \\
        \tc{1}    & $\text{30.46}_{\pm \text{2.13}}$ & $\text{13.61}_{\pm \text{0.64}}$ & $\text{10.00}_{\pm \text{2.67}}$ & $\text{5.87}_{\pm \text{0.80}}$ & $\text{8.97}_{\pm \text{2.43}}$ & $\text{6.45}_{\pm \text{0.91}}$ & $\text{14.46}_{\pm \text{2.93}}$ & $\text{8.13}_{\pm \text{0.93}}$ \\
        \tc{5}    & $\text{31.47}_{\pm \text{2.15}}$ & $\text{13.11}_{\pm \text{0.62}}$ & $\text{10.03}_{\pm \text{2.62}}$ & $\text{8.87}_{\pm \text{1.04}}$ & $\text{9.25}_{\pm \text{2.45}}$ & $\text{8.09}_{\pm \text{0.99}}$ & $\text{14.45}_{\pm \text{2.93}}$ & $\text{8.77}_{\pm \text{0.95}}$ \\
        \tc{10}   & $\text{31.41}_{\pm \text{2.13}}$ & $\text{12.68}_{\pm \text{0.59}}$ & $\text{10.05}_{\pm \text{2.63}}$ & $\text{9.74}_{\pm \text{1.06}}$ & $\text{9.31}_{\pm \text{2.46}}$ & $\text{8.37}_{\pm \text{1.00}}$ & $\text{14.23}_{\pm \text{2.90}}$ & $\text{8.89}_{\pm \text{0.97}}$ \\
        \midrule
        \qlk      & $\textbf{39.32}_{\pm \text{2.21}}$ & $\text{1.97}_{\pm \text{0.33}}$ & $\textbf{21.82}_{\pm \text{3.71}}$ & $\text{1.43}_{\pm \text{0.62}}$ & $\textbf{14.14}_{\pm \text{2.71}}$ & $\text{1.45}_{\pm \text{0.56}}$ & $\textbf{21.69}_{\pm \text{3.48}}$ & $\text{0.55}_{\pm \text{0.27}}$ \\
        \qlv      & $\text{26.43}_{\pm \text{2.01}}$ & $\textbf{15.24}_{\pm \text{0.68}}$ & $\text{10.27}_{\pm \text{2.69}}$ & $\textbf{10.63}_{\pm \text{1.03}}$ & $\text{8.61}_{\pm \text{2.35}}$ & $\textbf{9.26}_{\pm \text{1.00}}$ & $\text{11.65}_{\pm \text{2.58}}$ & $\textbf{9.36}_{\pm \text{0.99}}$ \\
        \bottomrule
      \end{tabular}
    \end{footnotesize}
  \end{center}
\end{table*}

\subsection{Definition of Query Lens}
\label{sec:feature-interpretation}

We finally propose two variants of \textbf{Query Lens (QL)} by configuring the three elements of the residual dynamics.

\paragraph{Value Variant (\qlv).}
We score tokens by transporting the value feature to the output endpoint with the full stream transition and reading out through the unembedding:
\begin{equation}
s_{\textsc{Value}}
\;=\;
U^{\top}\!\left[\prod_{k>l}\bigl(I + J^{k}\bigr)\right] v_i^{\,l}
\;\in\; \mathbb{R}^{|V|}.
\label{eq:s_value}
\end{equation}
We take the top-$k$ tokens under $s_{\textsc{Value}}$ as the tokens that the feature promotes at the output when activated, interpreting them as the feature's \emph{output-side} causal role.

\paragraph{Key Variant (\qlk).}
We score tokens by transporting the key feature to the input endpoint with the full stream transition and reading out in terms of token substitutions:
\[
s_{\textsc{Key}}^{\top}
\;=\;
(k_i^{\,l})^{\top}\!\left[\prod_{k\le l}\bigl(I + J^{k}\bigr)\right]\widehat{E}
\;\in\;
\mathbb{R}^{|V|}.
\]
We take the top-$k$ tokens under $s_{\textsc{Key}}$ as the tokens that most increase the feature activation, interpreting them as the feature's \emph{input-side} causal role. We describe how both scores are computed efficiently in Appendix~\ref{app:ql-compute}.

\begin{figure*}[t]
  \centering

  \includegraphics[width=0.38\textwidth]{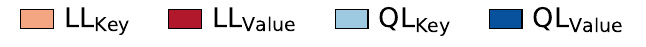}
  \vspace{0.05in}

  \begin{subfigure}[t]{0.245\textwidth}
    \centering
    \includegraphics[width=\linewidth]{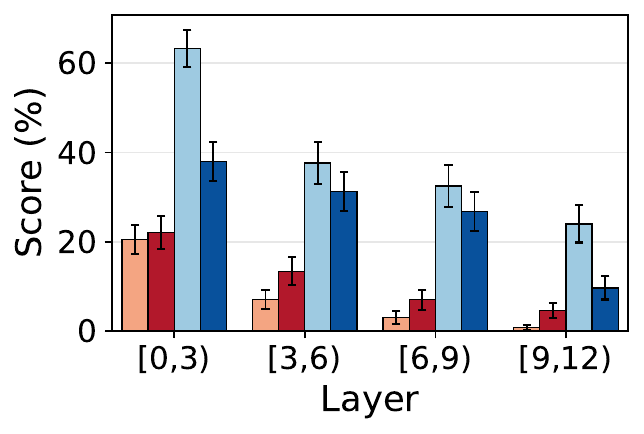}
    \caption{$I(T)$: GPT-2 Small}
    \label{fig:alt-gpt2-input}
  \end{subfigure}\hfill
  \begin{subfigure}[t]{0.245\textwidth}
    \centering
    \includegraphics[width=\linewidth]{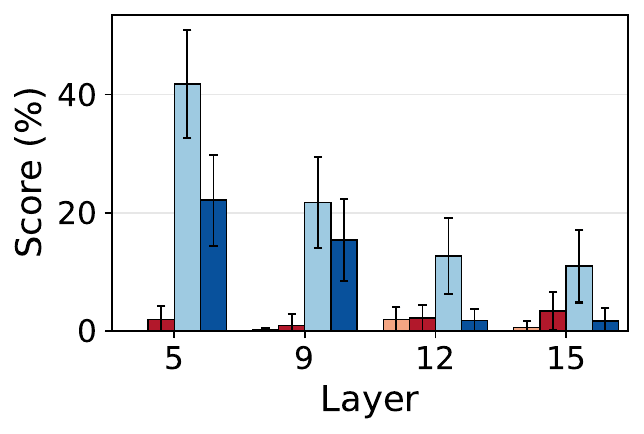}
    \caption{$I(T)$: Gemma-3-270M}
    \label{fig:alt-gemma270m-input}
  \end{subfigure}\hfill
  \begin{subfigure}[t]{0.245\textwidth}
    \centering
    \includegraphics[width=\linewidth]{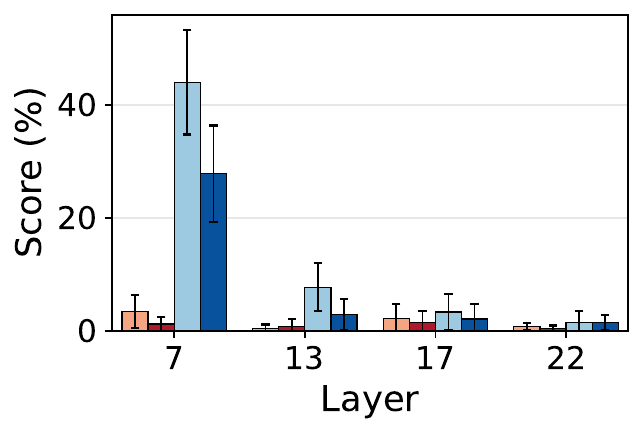}
    \caption{$I(T)$: Gemma-3-1B}
    \label{fig:alt-gemma1b-input}
  \end{subfigure}\hfill
  \begin{subfigure}[t]{0.245\textwidth}
    \centering
    \includegraphics[width=\linewidth]{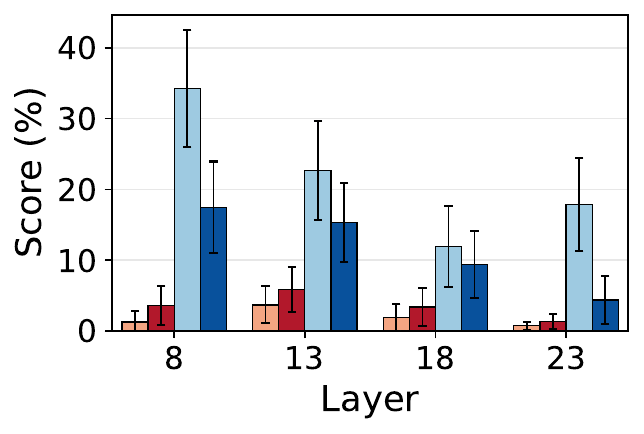}
    \caption{$I(T)$: Qwen-3-1.7B}
    \label{fig:alt-qwen-input}
  \end{subfigure}

  \vspace{0.05in}

  \begin{subfigure}[t]{0.245\textwidth}
    \centering
    \includegraphics[width=\linewidth]{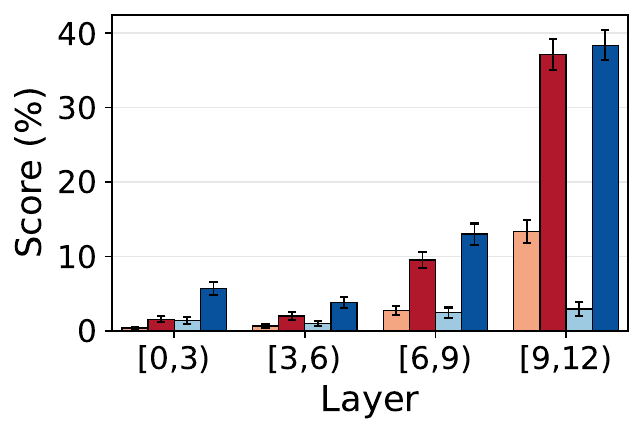}
    \caption{$O(T)$: GPT-2 Small}
    \label{fig:alt-gpt2-output}
  \end{subfigure}\hfill
  \begin{subfigure}[t]{0.245\textwidth}
    \centering
    \includegraphics[width=\linewidth]{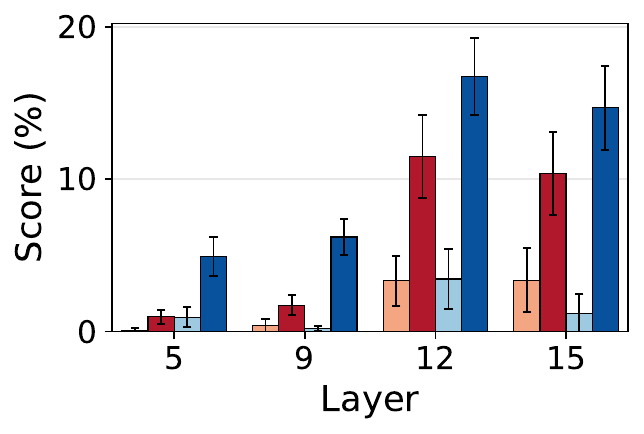}
    \caption{$O(T)$: Gemma-3-270M}
    \label{fig:alt-gemma270m-output}
  \end{subfigure}\hfill
  \begin{subfigure}[t]{0.245\textwidth}
    \centering
    \includegraphics[width=\linewidth]{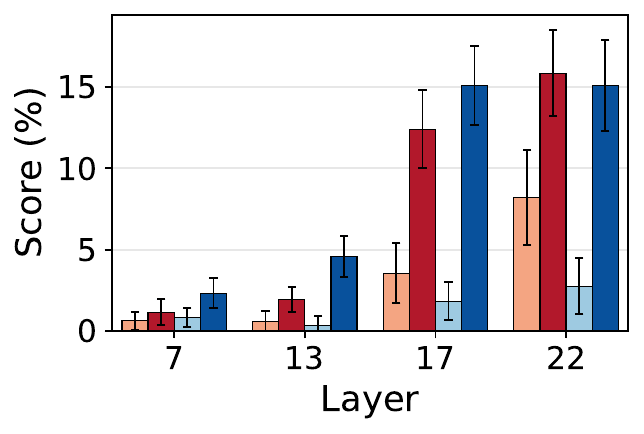}
    \caption{$O(T)$: Gemma-3-1B}
    \label{fig:alt-gemma1b-output}
  \end{subfigure}\hfill
  \begin{subfigure}[t]{0.245\textwidth}
    \centering
    \includegraphics[width=\linewidth]{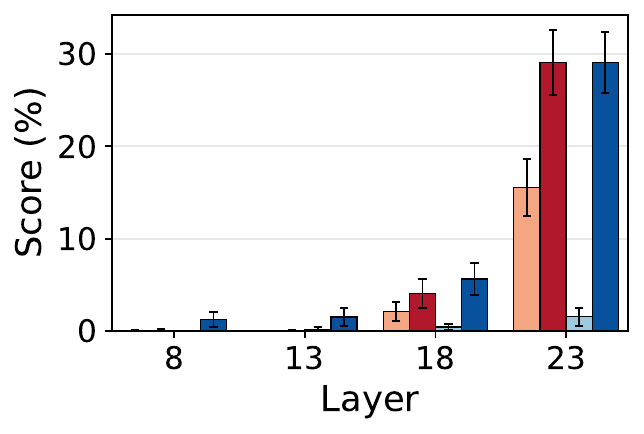}
    \caption{$O(T)$: Qwen-3-1.7B}
    \label{fig:alt-qwen-output}
  \end{subfigure}

  \caption{Input (top row) and output (bottom row) scores by layer group for four model/SAE settings, comparing the Logit Lens (\llk, \llv) and Query Lens (\qlk, \qlv) variants from Table~\ref{tab:main}. Each bar shows the per-method score for the indicated layer (or averaged within a layer group); error bars are 95\% confidence intervals.}
  \label{fig:layergroup-scores}
\end{figure*}

\section{Experiments}
\label{section:experiments}

In this section, we evaluate whether Query Lens better captures a feature's causal behavior from both the input and output sides, compared to baselines including Logit Lens.

\subsection{Experimental Configuration}
\label{subsec:exp-config}

\paragraph{Features.}
We analyze sparse dictionary features across four LLMs spanning three model families: GPT-2 Small \cite{radford2019language}, Gemma-3 (270M and 1B) \cite{gemmateam2025gemma3technicalreport}, and Qwen-3-1.7B \cite{qwen3technicalreport}.
For GPT-2 Small, we use the OpenAI Top-K SAE (32K)~\cite{gao2025scaling}.
For Gemma-3, we use Gemma Scope 2 JumpReLU SAEs (65K)~\cite{mcdougall2025gemma_scope_2}.
For Qwen-3-1.7B, we use the Qwen-Scope Top-K SAE (32K)~\cite{qwen_scope}.
Details of the specific model checkpoints and SAEs used in our experiments are listed in Appendix~\ref{app:model-sae-repos}.
For GPT-2 Small, we randomly sample 100 features per layer; for Gemma-3 and Qwen-3-1.7B, we randomly sample 100 features from each of four representative layers.
Additional results (different SAE widths, model sizes, and transcoders~\cite{dunefsky2024transcodersinterpretablellmfeature}) are reported in Appendix~\ref{app:additional-configurations}.

\paragraph{Baselines and Setup.}
We compare Query Lens against four baselines.
To enable a direct comparison, we first define the following \textbf{Logit Lens (LL)} counterparts:
\textbf{\llk}, which utilizes the key feature, the identity transition, and the embedding readout; and \textbf{\llv}, which employs the value feature, the identity transition, and the unembedding readout.
\textbf{Tuned Lens (TL)} \cite{belrose2025elicitinglatentpredictionstransformers} extends \llv\ by replacing the identity transition with a per-layer affine map $(A^{l}, b^{l})$ trained to transport intermediate hidden states toward the final-layer representation:
\[
s_{\textsc{TL}}
\;=\;
U^{\top} (A^{l}\, v_i^{\,l} + b^{l}).
\]
\textbf{Zero-Out (ZO)} and \textbf{Token Change (TC)} \cite{templeton2024scaling, gur-arieh-etal-2025-enhancing} are ablation-style baselines that take a finite difference of the logits between two operating points $(a^{-}, a^{+})$ on $y=f(a)$:
\[
s_{\textsc{ZO/TC}}
\;=\;
y\!\left(a_{\text{post}}^{l,i}=a^{+}\right) - y\!\left(a_{\text{post}}^{l,i}=a^{-}\right),
\]
with $(a^{-}, a^{+}) = (0,\, a_{\text{clean}})$ for ZO and $(a^{-}, a^{+}) = (a_{\text{clean}},\, a_{\text{clamp}})$ for TC, where $a_{\text{clean}}$ is the original value of $a$; we report TC at three clamp settings, $a_{\text{clamp}}\in\{1,5,10\}$ (denoted \tc{1}, \tc{5}, \tc{10}).

In modeling how the output depends on a feature's activation $a$ (or, on the key side, how the input drives $a$), each method diverges from QL as follows: 
LL and TL can be viewed as taking tangents of a simplified model whose stream transition is linearized (identity for LL and a learned affine map for TL); 
ZO and TC are secants between two operating points, scoring by the logit difference $y(a^{+}) - y(a^{-})$, while QL takes a tangent, not a secant, of the true rather than a linearized transition. 
Appendix~\ref{app:linearization-schematic} gives a geometric view.

For each method, we denote by $T$ the set of top-$k$ tokens it identifies for a given feature, and we fix $k=25$ throughout. 
Further implementation details are provided in Appendix~\ref{app:baseline-impl}.

\subsection{Quantitative Evaluation}
\label{subsection:quantitative}

\paragraph{Input Score.}
\label{subsection:input-side}
To test whether the tokens identified for each feature by a given method correspond to inputs that strongly activate the feature, we define the \emph{input score} as follows.
For each feature, we collect a set of sentences $S$ (with $|S|\ge 20$) where it exhibits non-zero activation on at least one token \cite{bills2023language,choi2024automatic,paulo2025automaticallyinterpretingmillionsfeatures}.
For each sentence in $S$, we select the most activated token and define $A$ as the resulting set across $S$.
The input score $I(T)$ is the fraction of tokens in $A$ appearing in $T$:
\[
I(T)
\;=\;
\frac{\left|\{t\in A \mid t\in T\}\right|}{|A|}.
\]
Intuitively, a high $I(T)$ means that the method's proposed tokens are themselves the ones that strongly activate the feature in natural inputs.

\paragraph{Output Score.}
\label{subsection:output-side}
To evaluate whether the tokens identified for each feature by a given method correspond to tokens that the feature causally promotes in generation, we define the \emph{output score} as follows.
We construct a set of $N=100$ neutral prefixes $\mathcal{P}=\{p_j\}_{j=1}^{N}$ (e.g., \emph{``Findings show that''}).
For each prefix $p\in\mathcal{P}$, we obtain two next-token distributions: a \emph{clean} distribution without intervention, and a \emph{steered} distribution in which we clamp the feature post-activation to a value $\alpha$ drawn from a set $\mathcal{A}$ of steering strengths.
For each token $t\in V$, let $\rho^{p}_{\text{clean}}(t)$ and $\rho^{p}_{\alpha}(t)\in[0,1]$ denote its rank percentile under the clean and steered next-token distributions.
The per-token rank-percentile delta induced by steering is
\[
\Delta\rho(t)
\;=\;
\frac{1}{|\mathcal{P}|}\sum_{p\in\mathcal{P}}
\frac{1}{|\mathcal{A}|}\sum_{\alpha\in\mathcal{A}}
\bigl(\rho^{p}_{\alpha}(t) - \rho^{p}_{\text{clean}}(t)\bigr),
\]
and we let $S$ be the top-$25$ tokens by $\Delta\rho$, i.e., those most promoted by steering.
The output score $O(T)$ is defined as the fraction of tokens in $S$ appearing in $T$:
\[
O(T)
\;=\;
\frac{\left|\{\,t\in S\mid t\in T\,\}\right|}{|S|}.
\]
Intuitively, a high $O(T)$ means that the method's proposed tokens are themselves the ones the feature promotes during generation.
See Appendix~\ref{app:eval-details} for further evaluation details.

\begin{figure*}[t]
  \centering
  \includegraphics[width=0.95\linewidth]{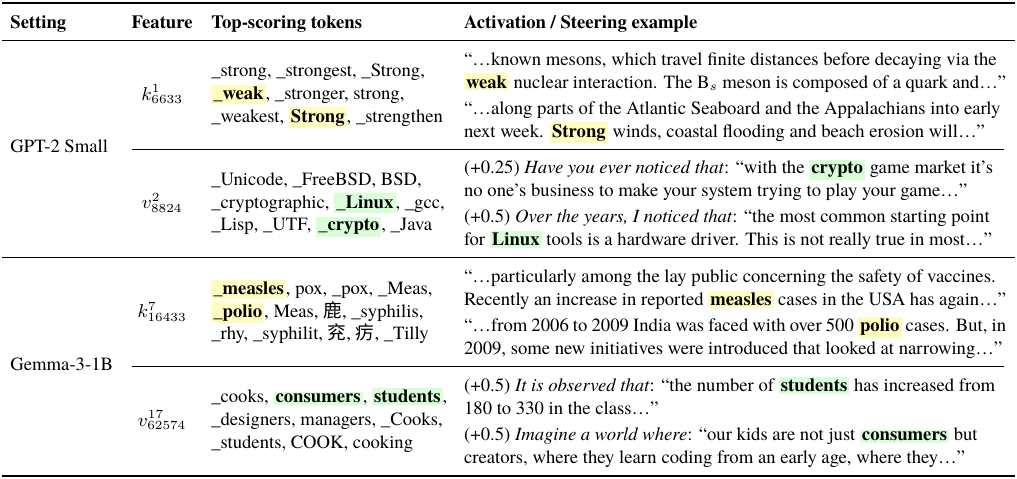}
  \caption{Qualitative examples showing that \qlk\ tokens (\hky{yellow}) match the inputs that activate the feature (i.e., high $I(T)$) and \qlv\ tokens (\hkg{green}) match the outputs promoted by steering (i.e., high $O(T)$), on GPT-2 Small and Gemma-3-1B. Steering examples are prefixed with the steering factor $\alpha$ and the base prompt.}
  \label{tab:qual-causal}
\end{figure*}

\paragraph{Results.}
\label{subsection:results-quant}
Tables~\ref{tab:main} report input and output scores averaged across features for each model and SAE. Across all configurations, \textbf{\qlk} achieves the best $I(T)$ and \textbf{\qlv} achieves the best $O(T)$. 
Two factors explain this pattern.
(1) \textbf{Faithfulness:} On each side, the QL variant outperforms its LL counterpart, confirming that modeling indirect effects contributes to more faithful interpretations.
(2) \textbf{Completeness:} \qlk\ achieves the highest $I(T)$, but trails on $O(T)$, while \qlv\ shows the opposite, and no single variant is best on both sides. 
This confirms that designing dedicated key and value variants is effective for a complete interpretation of a feature's causal roles. 
Figure~\ref{tab:qual-causal} illustrates examples of QL's tokens and the actual model behavior they explain.

Figure~\ref{fig:layergroup-scores} breaks both scores down by layer group. 
We observe two patterns.
First, both $I(T)$ and $O(T)$ decrease as the feature moves farther from its corresponding endpoint (later layers for $I(T)$, earlier layers for $O(T)$). 
Longer stream transitions involve more intervening modules, producing larger indirect effects that form a bottleneck for predicting the feature's causal effect.
Second, QL's relative gain over LL widens as the feature moves farther from the endpoint. 
This confirms QL's design: it captures the indirect effects that LL misses, and its contribution scales with the size of those effects.
Appendix~\ref{app:A} provides an analysis of how the impact of indirect effects varies with layer depth.

For TL, it underperforms QL across all settings. 
We attribute this failure to a distributional mismatch: the learned map is trained on full residual stream vectors $h^l$, which are dense combinations of many concurrently active features. 
A single SAE feature vector $v_i^l$ has a different norm and structure, so the map does not generalize reliably to such inputs. 
For TC, the output score can be competitive at certain clamp values, but the best clamp varies across settings, and within a single setting, the score remains sensitive to the clamp choice.

\begin{table}[t]
  \caption{Interpretability scores (mean $\pm$ 95\% CI half-width) of token-set explanations.
  Within each Key (\llk vs. \qlk) and Value (\llv vs. \qlv) pair, the larger value is in \textbf{bold}.}
  \label{tab:interp-score}
  \centering
  \begin{small}
  \renewcommand{\arraystretch}{1.15}
    \setlength{\tabcolsep}{4pt}
    \begin{tabular}{l
        >{\columncolor{inputbg}}c
        c
        >{\columncolor{inputbg}}c
        c}
      \toprule
      Setting & \llk & \qlk & \llv & \qlv \\
      \midrule
      GPT-2 Small
      & $\text{3.03}_{\pm \text{0.12}}$
      & $\textbf{4.02}_{\pm \text{0.14}}$
      & $\text{4.87}_{\pm \text{0.15}}$
      & $\textbf{5.95}_{\pm \text{0.16}}$ \\
      Gemma-3-270M
      & $\text{4.00}_{\pm \text{0.27}}$
      & $\textbf{5.44}_{\pm \text{0.24}}$
      & $\text{4.10}_{\pm \text{0.27}}$
      & $\textbf{5.34}_{\pm \text{0.28}}$ \\
      Gemma-3-1B
      & $\text{4.00}_{\pm \text{0.29}}$
      & $\textbf{4.86}_{\pm \text{0.24}}$
      & $\text{4.58}_{\pm \text{0.29}}$
      & $\textbf{5.30}_{\pm \text{0.29}}$ \\
      Qwen-3-1.7B
      & $\text{3.47}_{\pm \text{0.24}}$
      & $\textbf{4.76}_{\pm \text{0.28}}$
      & $\text{5.06}_{\pm \text{0.30}}$
      & $\textbf{6.96}_{\pm \text{0.30}}$ \\
      \bottomrule
    \end{tabular}
  \end{small}
\end{table}

\begin{figure}[t]
  \centering
  \begin{subfigure}[t]{\linewidth}
    \centering
    \includegraphics[width=0.9\linewidth]{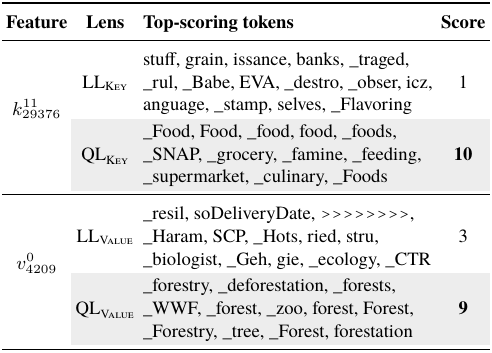}
    \caption{GPT-2 Small}
    \label{tab:qual-semantic-gpt2}
  \end{subfigure}

  \vspace{0.6em}
  \begin{subfigure}[t]{\linewidth}
    \centering
    \includegraphics[width=0.9\linewidth]{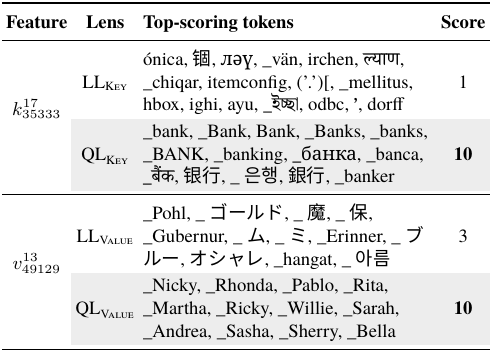}
    \caption{Gemma-3-1B}
    \label{tab:qual-semantic-gemma1b}
  \end{subfigure}
  \caption{Qualitative examples of Logit Lens and Query Lens top tokens with their interpretability scores on GPT-2 Small and Gemma-3-1B. The higher score per block is shown in \textbf{bold}.}
  \label{tab:qual-semantic}
\end{figure}

\subsection{Qualitative Evaluation}
\label{sec:qual-eval}
We evaluate whether token sets identified by Query Lens form more human-interpretable semantic units than those from Logit Lens.
We prompt GPT-5-nano \cite{2025gpt5} to assign an interpretability score ($0$–$10$) to each feature's top-$k$ tokens based on how concentrated they are around a single coherent theme; see Appendix~\ref{app:G} for the full prompt.

Table~\ref{tab:interp-score} shows that Query Lens consistently achieves higher interpretability scores than Logit Lens, indicating that its token sets converge more reliably to coherent semantic descriptions.
Figure~\ref{tab:qual-semantic} illustrates this pattern with examples: Logit Lens top-tokens are dominated by tokenizer fragments while Query Lens top-tokens form a single recognizable concept, with interpretability scores rising accordingly.

\section{Subspace Channel Hypothesis}
\label{sec:subspace-channel}
In this section, we conduct a focused analysis of module-mediated effects captured by Query Lens.

\paragraph{Intuition.}
Beyond feature interpretation, Query Lens affords the following \emph{query--response} view: along the residual stream, a feature acts as a \emph{query} to each downstream module, which writes back its \emph{response}.
To make this concrete, we simplify the stream transition by keeping only the identity and first-order Jacobian terms:
\[
s_{\textsc{Value}}
\;\approx\;
U^{\top}\!\left(v_i^{\,l} \;+\; \sum_{k>l} J^{k} v_i^{\,l}\right).
\]
Each Jacobian term then represents a single-hop interaction between the feature and a downstream module: the value feature $v_i^{\,l}$ is the \emph{query}, while each Jacobian--vector product (JVP) $r_{i}^{\,l\rightarrow k} = J^{k} v_i^{\,l} \in \mathbb{R}^{d_{\mathrm{m}}}$ is the \emph{module response} produced by the module at layer $k$.

\paragraph{Hypothesis.}
Motivated by the observation that modules respond differently to the same feature, we hypothesize that downstream modules read a query $v_i^{\,l}$ \emph{selectively} through module-specific low-dimensional subspaces.
In other words, for a pair of layers $(l,k)$ with $k>l$, we posit that the response at layer $k$ is primarily determined by the projection of $v_i^{\,l}$ onto a subspace associated with the downstream module, rather than by the full feature direction.
We refer to this module-specific subspace as a \emph{channel}, and term this the \textbf{Subspace Channel Hypothesis}.

\begin{figure}[t]
  \centering
  \begin{subfigure}[t]{\linewidth}
    \centering
    \includegraphics[width=0.95\linewidth]{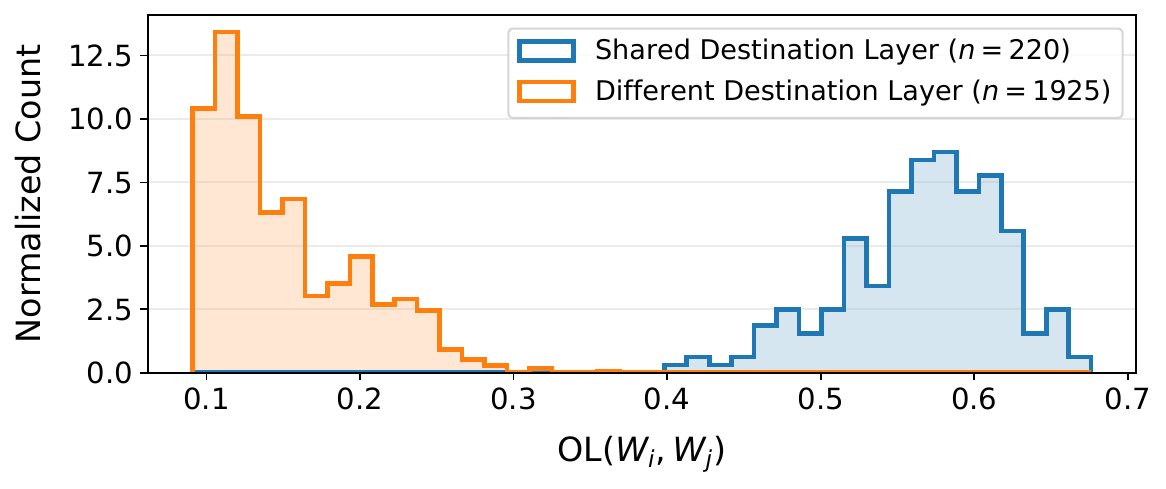}
    \caption{Distribution of $\mathrm{OL}$, grouped by whether the pair shares the same destination layer. $n$ is the number of pairs per group.}
    \label{fig:overlap-distribution}
  \end{subfigure}

  \vspace{0.6em}
  \begin{subfigure}[t]{\linewidth}
    \centering
    \includegraphics[width=0.95\linewidth]{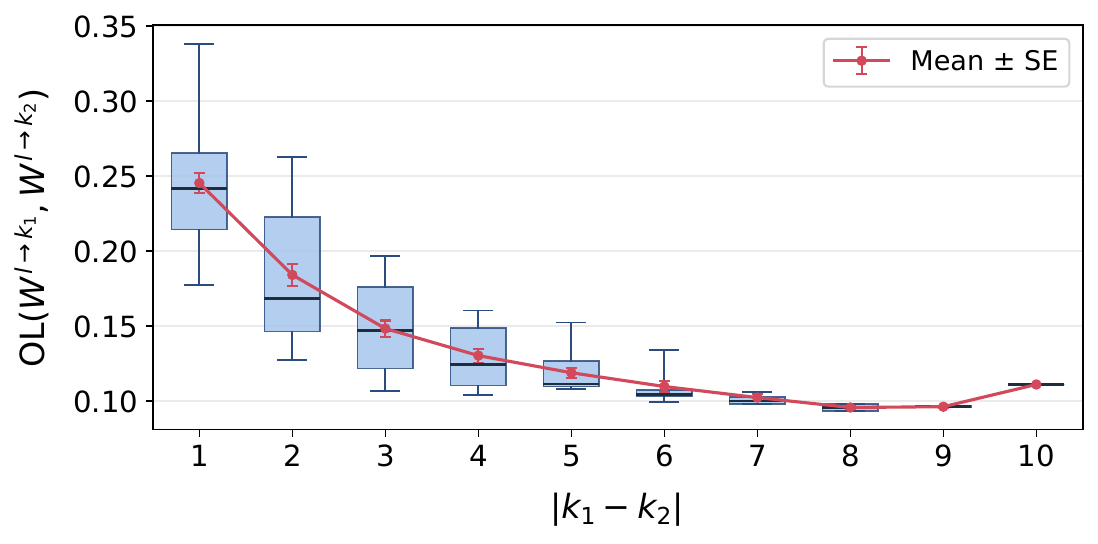}
    \caption{$\mathrm{OL}$ as a function of the distance $|k_1-k_2|$ between destination layers, for pairs that share a source layer.}
    \label{fig:overlap-vs-distance}
  \end{subfigure}
  \caption{Overlap statistics for the learned maps $\{W^{l\rightarrow k}\}$. (a) Pairs sharing a destination layer cluster at higher $\mathrm{OL}$ than those with different destinations. (b) Among pairs sharing a source layer, overlap decays with the distance between destination layers.}
  \label{fig:overlap-figs}
\end{figure}

\paragraph{Experiments.}
To verify our hypothesis, we conduct an experiment according to the following procedure:

\begin{enumerate}[itemsep=0.35em, topsep=0.35em, leftmargin=*]
    \item \textbf{Sample queries.}
    For each layer $l$ of GPT-2 Small ($L=12$), we sample $N=1000$ value features $\{v_i^{\,l}\}_{i=1}^{N}$.

    \item \textbf{Compute module responses.}
    For each sampled feature $v_i^{\,l}$, we compute its module response $r_{i}^{\,l\rightarrow k}$ at all downstream layers $k>l$.
    Across the model, there are $\binom{L}{2}=66$ layer pairs $(l,k)$, and for each pair we obtain $N$ $(\text{query},\text{response})$ pairs by evaluating the responses induced by the $N$ sampled queries at layer $l$.

    \item \textbf{Learn low-rank maps from queries to responses.}
    For each layer pair $(l,k)$, we fit a linear map from queries at layer $l$ to responses at layer $k$:
    \[
    r_{i}^{\,l\rightarrow k}
    \;\approx\;
    W^{l\rightarrow k} v_i^{\,l},
    \qquad
    W^{l\rightarrow k}\in\mathbb{R}^{d_{\mathrm{m}}\times d_{\mathrm{m}}}.
    \]
    To enforce low dimensionality, each map is constrained to be low-rank via a LoRA-style factorization, $W^{l\rightarrow k}=B^{l\rightarrow k}A^{l\rightarrow k}$.
    We use $A^{l\rightarrow k}\in\mathbb{R}^{r\times d_{\mathrm{m}}}$ and $B^{l\rightarrow k}\in\mathbb{R}^{d_{\mathrm{m}}\times r}$ with $r=d_{\mathrm{m}}/n_{\mathrm{layers}}=64$. This factorization restricts the map to select an effective $r$-dimensional basis that best predicts downstream JVPs from the query feature directions. Training details are in Appendix~\ref{app:D.1}.

    \item \textbf{Compare learned maps.}
    We consider all pairs $\{(W_i, W_j)\}_{1 \le i < j \le |\mathcal{W}|}$ from the learned matrices $\mathcal{W}=\{W^{l\rightarrow k}\}$, and measure the overlap between their column spaces, denoted by $\mathrm{OL}$ (short for \emph{overlap}).
     Let $Q_i\in\mathbb{R}^{d_{\mathrm{m}}\times r}$ and $Q_j\in\mathbb{R}^{d_{\mathrm{m}}\times r}$ be orthonormal basis matrices for the column spaces of $W_i$ and $W_j$. 
     We define
    \[
    \mathrm{OL}(W_i,W_j)
    \;=\;
    \frac{1}{r}\sum_{p=1}^{r}\cos^2\theta_p
    \;=\;
    \frac{1}{r}\left\|Q_i^\top Q_j\right\|_F^2,
    \]
    where $\{\theta_p\}_{p=1}^{r}$ are the principal angles between the two $r$-dimensional subspaces. 
    This measures the degree of sharing channels between two distinct feature readings.
\end{enumerate}

\paragraph{Result.} Figure~\ref{fig:overlap-distribution} summarizes pairwise subspace overlap among the learned maps $\{W^{l\rightarrow k}\}$, grouped by whether two maps share same destination layer $k$. 
In particular, pairs that share the target layer (e.g., $W^{0\rightarrow5}$ and $W^{2\rightarrow5}$) concentrate at substantially higher $\mathrm{OL}$ values, whereas pairs with different targets are skewed toward low overlap. 
This indicates that the column space of $W^{l\rightarrow k}$---low-dimensional \emph{channels} for feature reading---is consistent across different source layers $l$ for a fixed $k$. 
Conversely, when the consuming module changes (i.e., different $k$), the channels become non-overlapping, suggesting that each module reads queries through its own subspace.

We further examine how channel similarity depends on destination distance (Fig.~\ref{fig:overlap-vs-distance}). For a fixed source layer $l$, we calculate overlap between two maps, $\mathrm{OL}(W^{l\rightarrow k_1},W^{l\rightarrow k_2})$, and aggregate the results by the distance between target layers, i.e., $|k_1- k_2|$. 
We find that pairwise overlap decays as the distance between destination layers grows. 
This indicates that the channel is shared more between nearby modules and becomes increasingly distinct with distance.

\section{Related Work}

\paragraph{Feature Interpretation.}
Approaches to interpreting SAE features broadly fall into two main lines.
\textbf{Activation-based} methods run the model on large corpora and use the inputs that strongly activate a feature to summarize its meaning \cite{bills2023language,bricken2023monosemanticity,choi2024automatic,paulo2025automaticallyinterpretingmillionsfeatures,templeton2024scaling}.
\textbf{Parameter-based} methods, in contrast, read meaning directly from learned weights, most commonly by projecting SAE decoder vectors into vocabulary space via Logit Lens \cite{nostalgebraist2020interpreting,bloom2024understandingfeatureslogitlens}.
The two lines differ not only in computational style (corpus collection vs.\ static projection) but also in what they observe: the former reveals which inputs activate a feature, while the latter reveals which outputs it promotes \cite{gur-arieh-etal-2025-enhancing,arad-etal-2025-saes}.
Parameter-based methods, however, are not intrinsically restricted to the output side, as the key--value memory view of MLPs shows \cite{geva-etal-2021-transformer,geva-etal-2022-transformer,dar-etal-2023-analyzing}.
Yet SAE feature interpretation has kept this split tied to methodology (activation-based for input, parameter-based for output), and Query Lens closes this gap by deriving both interpretations from analogous formulations over the SAE's encoder and decoder directions.

Other extensions of the parameter-based work build on Logit Lens with different goals: \citet{belrose2025elicitinglatentpredictionstransformers} improve calibration with per-layer affine maps (\emph{Tuned Lens}); \citet{katz-etal-2024-backward} project gradients to interpret learning dynamics (\emph{Backward Lens}); \citet{hernandez2024linearity} decode relation-specific attributes via linear relational embeddings (\emph{Attribute Lens}); and \citet{pal-etal-2023-future} anticipate tokens several positions ahead from a single hidden state (\emph{Future Lens}).
Query Lens directs this extension toward faithful interpretation of SAE features by accounting for the indirect effects.

\paragraph{Layer Communication.}
\citet{elhage2021mathematical} conceptualized the residual stream as a high-dimensional \emph{communication channel}: each layer reads from and writes to low-rank subspaces of the stream, and the effective interaction between any two layers is captured by their \emph{virtual weights}.
\citet{merullo2024talking} concretized this view at the level of attention heads, applying SVD to their QK/OV weight matrices to identify low-rank channels through which earlier heads pass information to later ones.
The Subspace Channel Hypothesis extends this picture from attention heads to layers (attention and MLP), taking a functional view of how each downstream layer responds to an SAE feature.

\section{Conclusion}
We introduce \textbf{Query Lens}, a parameter-based method for SAE feature interpretation that improves Logit Lens along two dimensions: it spans both encoder and decoder directions to characterize input- and output-side causality, and it models the indirect, module-mediated effects of features in the residual stream.
Across settings, Query Lens produces more interpretable token signatures than Logit Lens and prior baselines.
Building on this framework, we propose the \textbf{Subspace Channel Hypothesis} and provide evidence that downstream modules read features through layer-specific low-rank subspaces.
An open question that Query Lens raises is what function querying achieves: why features develop in such a form, and how it shapes model behavior. The discovered channels also open the door to practical applications such as model editing \cite{meng2023locatingeditingfactualassociations}.





\section*{Impact Statement}
This paper presents work whose goal is to advance the field of mechanistic interpretability. While progress in machine learning can have broad societal consequences, we believe that research aimed at improving understanding of ML models is unlikely to introduce additional harms beyond those already associated with the underlying models, and instead provides tools that can help identify, anticipate, and avert such harms.

\section*{Acknowledgements}
This work was supported by Institute of Information \& Communications Technology Planning \& Evaluation (IITP) grant funded by the Korea government (MSIT) (No.\ RS-2020-II201373, Artificial Intelligence Graduate School Program (Hanyang University)). This work was supported by Institute of Information \& Communications Technology Planning \& Evaluation (IITP) under the artificial intelligence semiconductor support program to nurture the best talents (IITP-(2026)-RS-2023-00253914) grant funded by the Korea government (MSIT). This work was supported by the National Research Foundation of Korea (NRF) grant funded by the Korea government (MSIT) (RS-2025-00558151).


\bibliography{example_paper}
\bibliographystyle{icml2026}

\newpage
\appendix
\onecolumn
\section{Models and SAEs}
\label{app:model-sae-repos}

Table~\ref{tab:hf-repos} lists the Hugging Face repositories for the LLMs and sparse autoencoders used in our main experiments (Section~\ref{section:experiments}).

\begin{table}[h]
\centering
\caption{Hugging Face repositories for the four model/SAE pairs evaluated in Table~\ref{tab:main}.}
\label{tab:hf-repos}
\small
\begin{tabular}{lll}
\toprule
Setting & Model Repository & SAE Repository \\
\midrule
GPT-2 Small  & \href{https://huggingface.co/openai-community/gpt2}{\texttt{openai-community/gpt2}}
             & \href{https://huggingface.co/jbloom/GPT2-Small-OAI-v5-32k-resid-post-SAEs}{\texttt{jbloom/GPT2-Small-OAI-v5-32k-resid-post-SAEs}} \\
Gemma-3-270M & \href{https://huggingface.co/google/gemma-3-270m}{\texttt{google/gemma-3-270m}}
             & \href{https://huggingface.co/google/gemma-scope-2-270m-pt}{\texttt{google/gemma-scope-2-270m-pt/resid\_post}} (l0\_medium) \\
Gemma-3-1B   & \href{https://huggingface.co/google/gemma-3-1b-pt}{\texttt{google/gemma-3-1b-pt}}
             & \href{https://huggingface.co/google/gemma-scope-2-1b-pt}{\texttt{google/gemma-scope-2-1b-pt/resid\_post}} (l0\_medium) \\
Qwen-3-1.7B  & \href{https://huggingface.co/Qwen/Qwen3-1.7B-Base}{\texttt{Qwen/Qwen3-1.7B-Base}}
             & \href{https://huggingface.co/Qwen/SAE-Res-Qwen3-1.7B-Base-W32K-L0_100}{\texttt{Qwen/SAE-Res-Qwen3-1.7B-Base-W32K-L0\_100}} \\
\bottomrule
\end{tabular}
\end{table}

Table~\ref{tab:hf-repos-additional} lists the repositories for the additional configurations evaluated in Appendix~\ref{app:additional-configurations}: the SAE settings (Table~\ref{tab:additional-sae}) and the Qwen-3 transcoders (Table~\ref{tab:additional-transcoder}).

\begin{table}[h]
\centering
\caption{Hugging Face repositories for the additional SAE configurations (Table~\ref{tab:additional-sae}) and transcoder configurations (Table~\ref{tab:additional-transcoder}).}
\label{tab:hf-repos-additional}
\small
\begin{tabular}{lll}
\toprule
Setting & Model Repository & Dictionary Repository \\
\midrule
GPT-2 Small (128K) & \href{https://huggingface.co/openai-community/gpt2}{\texttt{openai-community/gpt2}}
  & \href{https://huggingface.co/jbloom/GPT2-Small-OAI-v5-128k-resid-post-SAEs}{\texttt{jbloom/GPT2-Small-OAI-v5-128k-resid-post-SAEs}} \\
Gemma-3-270M (16K) & \href{https://huggingface.co/google/gemma-3-270m}{\texttt{google/gemma-3-270m}}
  & \href{https://huggingface.co/google/gemma-scope-2-270m-pt}{\texttt{google/gemma-scope-2-270m-pt/resid\_post}} (l0\_medium) \\
Gemma-3-1B (16K) & \href{https://huggingface.co/google/gemma-3-1b-pt}{\texttt{google/gemma-3-1b-pt}}
  & \href{https://huggingface.co/google/gemma-scope-2-1b-pt}{\texttt{google/gemma-scope-2-1b-pt/resid\_post}} (l0\_medium) \\
Gemma-3-4B (16K) & \href{https://huggingface.co/google/gemma-3-4b-pt}{\texttt{google/gemma-3-4b-pt}}
  & \href{https://huggingface.co/google/gemma-scope-2-4b-pt}{\texttt{google/gemma-scope-2-4b-pt/resid\_post}} (l0\_medium) \\
Gemma-3-4B (65K) & \href{https://huggingface.co/google/gemma-3-4b-pt}{\texttt{google/gemma-3-4b-pt}}
  & \href{https://huggingface.co/google/gemma-scope-2-4b-pt}{\texttt{google/gemma-scope-2-4b-pt/resid\_post}} (l0\_medium) \\
\midrule
Qwen-3-0.6B & \href{https://huggingface.co/Qwen/Qwen3-0.6B}{\texttt{Qwen/Qwen3-0.6B}}
  & \href{https://huggingface.co/mwhanna/qwen3-0.6b-transcoders-lowl0}{\texttt{mwhanna/qwen3-0.6b-transcoders-lowl0}} \\
Qwen-3-1.7B & \href{https://huggingface.co/Qwen/Qwen3-1.7B}{\texttt{Qwen/Qwen3-1.7B}}
  & \href{https://huggingface.co/mwhanna/qwen3-1.7b-transcoders-lowl0}{\texttt{mwhanna/qwen3-1.7b-transcoders-lowl0}} \\
Qwen-3-4B & \href{https://huggingface.co/Qwen/Qwen3-4B}{\texttt{Qwen/Qwen3-4B}}
  & \href{https://huggingface.co/mwhanna/qwen3-4b-transcoders}{\texttt{mwhanna/qwen3-4b-transcoders}} \\
\bottomrule
\end{tabular}
\end{table}

\section{Geometric Comparison of Methods}
\label{app:linearization-schematic}

\begin{figure}[ht]
  \centering
  \begin{subfigure}[t]{0.48\linewidth}
    \centering
    \includegraphics[width=\linewidth]{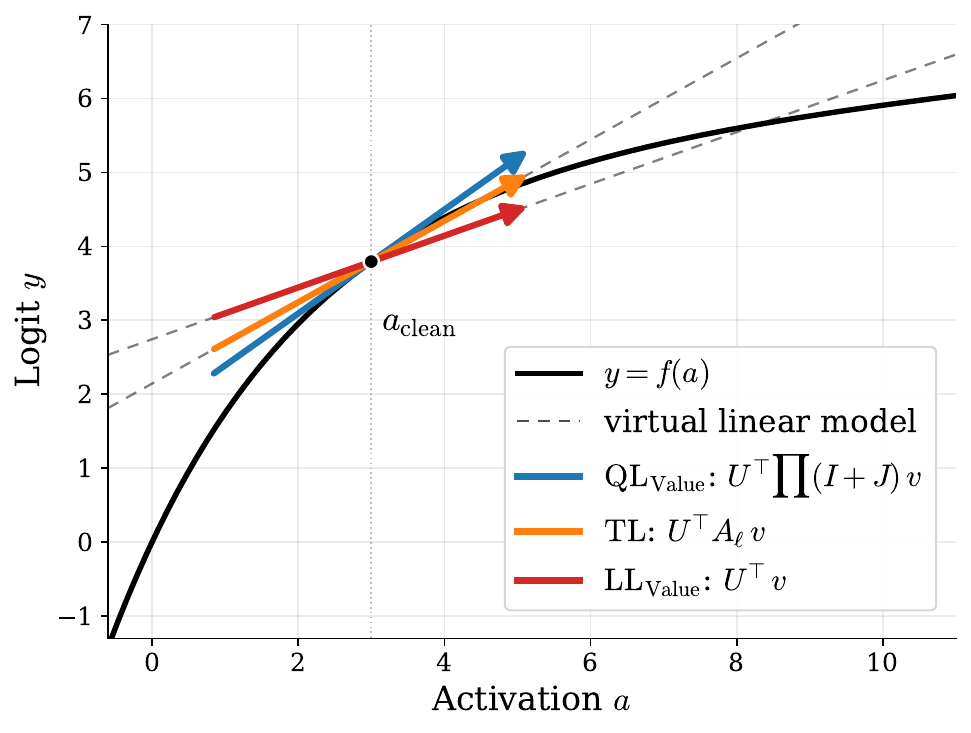}
    \caption{LL and TL are tangents of simplified linear surrogate models; QL is the local tangent of the true $f$ at $a_{\mathrm{clean}}$.}
    \label{fig:linearization-lens}
  \end{subfigure}\hfill
  \begin{subfigure}[t]{0.48\linewidth}
    \centering
    \includegraphics[width=\linewidth]{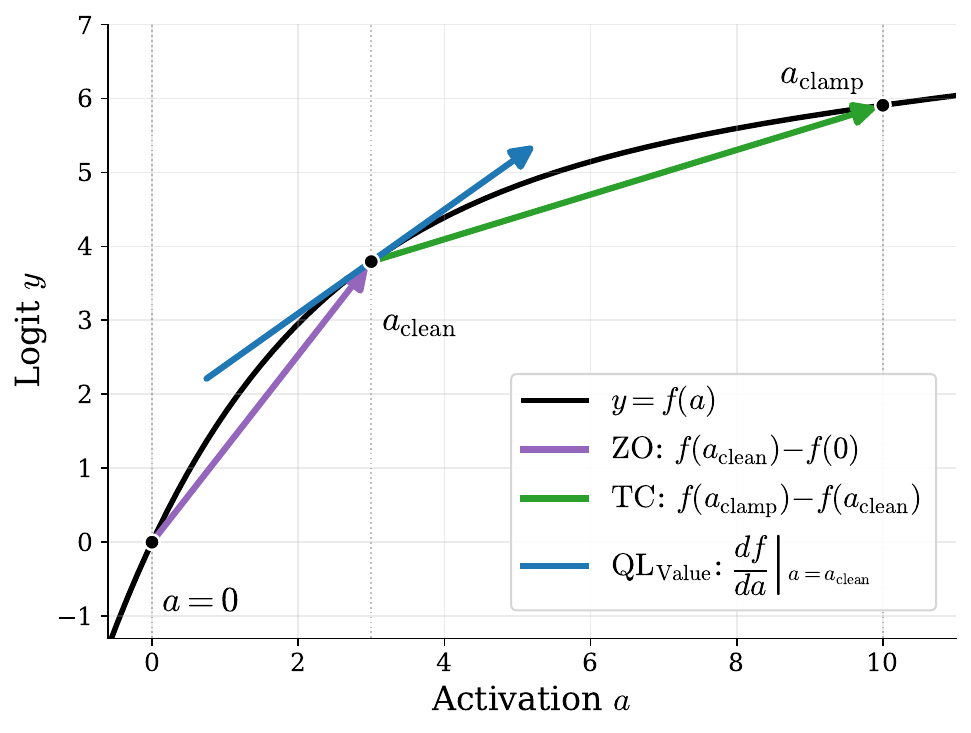}
    \caption{Query Lens vs.\ ablation-style surrogates: ZO and TC are secants anchored at $a=0$ and $a=a_{\mathrm{clamp}}$ respectively.}
    \label{fig:linearization-ablation}
  \end{subfigure}
  \caption{Schematic comparison of feature readouts as approximations to the true mapping $y=f(a)$ from feature activation $a$ to logit $y$. Additive constants are omitted in legend formulas.}
  \label{fig:linearization-schematic}
\end{figure}

Query Lens reads off the local tangent of the true activation-to-logit map $y=f(a)$ at the clean operating point $a_{\mathrm{clean}}$, with slope $f'(a_{\mathrm{clean}}) = U^{\top}\!\left[\prod_{k>l}(I + J^{k})\right] v_i^{\,l}$. Each baseline can be seen as diverging from this tangent in one of two ways, which we visualize in the two panels of Figure~\ref{fig:linearization-schematic}: (a) taking the tangent of a \emph{simplified, linearized} model rather than the true $f$, or (b) replacing the tangent with a \emph{secant} between two operating points.

\paragraph{(a) Tangent of a linearized model (LL, TL).}
Logit Lens and Tuned Lens take the slope at $a_{\mathrm{clean}}$ of a \emph{virtual linear model} obtained by simplifying the downstream stream transition, rather than of the true $f$ (Figure~\ref{fig:linearization-lens}). LL removes every downstream path that re-enters a subsequent module, leaving only the residual connections that carry the value vector directly to the output; the resulting model is linear in $a$, so its tangent is the constant slope $U^{\top} v_i^{\,l}$. TL is less severe: it collapses the downstream stream transition into a single learned per-layer affine map $W^{l}$, again yielding a linear model with constant slope $U^{\top} W^{l} v_i^{\,l}$. Both slopes are parameter-only and blind to the operating point, because the simplified model discards exactly the input-dependent indirect effects that QL retains.

\paragraph{(b) Secant instead of tangent (ZO, TC).}
Zero-Out and Token Change keep the true $f$ but replace the analytic tangent with a \emph{secant} between two operating points (Figure~\ref{fig:linearization-ablation}). ZO sets the activation to $0$ and reads off $s_{\textsc{ZO}} = f(a_{\mathrm{clean}}) - f(0)$, the secant of $f$ between $a=0$ and $a=a_{\mathrm{clean}}$. TC instead clamps the activation to a fixed value and measures $s_{\textsc{TC}} = f(a_{\mathrm{clamp}}) - f(a_{\mathrm{clean}})$, the secant between $a=a_{\mathrm{clean}}$ and $a=a_{\mathrm{clamp}}$. Unlike QL's tangent, a secant depends on the choice of the second point and averages the behavior of $f$ over a finite interval rather than at $a_{\mathrm{clean}}$ itself.

Across both axes, QL is the only surrogate that is simultaneously a tangent rather than a secant and taken on the true model rather than a simplified one, evaluated at the actual operating point.

\section{Baseline Implementation Details}
\label{app:baseline-impl}

\paragraph{Hidden-state Sampling for QL, ZO, and TC.}
QL, ZO, and TC all require evaluating the model at a clean operating point $a_{\text{clean}}$. We obtain these hidden states by running each model on a fixed corpus of $1024\times 32$ tokens drawn from The Pile~\cite{gao2020pile800gbdatasetdiverse}. As shown in Appendix~\ref{app:B2}, the token sets predicted by QL are largely invariant to this sample choice.

\paragraph{Tuned Lens.}
For GPT-2 Small, we use the pretrained lens released with the \texttt{tuned-lens} library~\cite{belrose2025elicitinglatentpredictionstransformers}.
For the remaining models, we train our own lens with the same library.
Each lens is a stack of per-layer linear translators that minimize the per-token KL divergence between the lens prediction at that layer and the (detached) final-layer next-token distribution, trained on the \texttt{wikitext-103-raw-v1}\footnote{\url{https://huggingface.co/datasets/Salesforce/wikitext}} validation split.
We optimize with SGD (Nesterov momentum $0.9$, weight decay $10^{-3}$) under a linear learning-rate decay to zero with no warmup, using a base learning rate of $0.01$ for $2000$ steps, batch size $1$, maximum sequence length $2048$, and \texttt{bfloat16} precision.
Table~\ref{tab:tuned-lens-eval} reports the resulting lens quality on held-out \texttt{wikitext-103} text: every lens substantially lowers the average per-token KL to the final-layer distribution relative to the logit lens (by 42--79\%).

\begin{table}[ht]
  \caption{Tuned lens quality, measured as the per-token KL divergence to the final-layer distribution, averaged over layers ($\downarrow$ lower is better); the parenthetical is the reduction relative to the logit lens. $^\dagger$From the \texttt{tuned-lens} library; the others we train ourselves.}
  \label{tab:tuned-lens-eval}
  \centering
  \begin{small}
    \begin{tabular}{lccc}
      \toprule
      Model & Layers & Logit KL ($\downarrow$) & Tuned KL ($\downarrow$) \\
      \midrule
      GPT-2 Small$^\dagger$ & 12 & 4.96  & 2.87 ($-42\%$) \\
      Gemma-3-270M         & 18 & 19.29 & 4.67 ($-76\%$) \\
      Gemma-3-1B           & 26 & 10.46 & 3.36 ($-68\%$) \\
      Gemma-3-4B           & 34 & 16.78 & 3.49 ($-79\%$) \\
      Qwen-3-0.6B          & 28 & 6.29  & 2.34 ($-63\%$) \\
      Qwen-3-1.7B          & 28 & 8.37  & 2.25 ($-73\%$) \\
      Qwen-3-1.7B-Base     & 28 & 8.74  & 2.12 ($-76\%$) \\
      Qwen-3-4B            & 36 & 8.62  & 2.35 ($-73\%$) \\
      \bottomrule
    \end{tabular}
  \end{small}
\end{table}

\section{Additional Configurations}
\label{app:additional-configurations}

In this section, we report Query Lens results on configurations omitted from the main text (Section~\ref{section:experiments}): additional SAE widths and model sizes (Section~\ref{app:additional-sae}), and transcoders applied to Qwen-3 models (Section~\ref{app:additional-transcoder}).

\subsection{Additional SAEs}
\label{app:additional-sae}

We evaluate Query Lens on five additional model/SAE configurations: GPT-2 Small with the 128K-width OpenAI Top-K SAE~\cite{gao2025scaling}, Gemma-3-270M and Gemma-3-1B with the 16K-width Gemma Scope~2 JumpReLU SAEs~\cite{mcdougall2025gemma_scope_2}, and Gemma-3-4B with both 16K and 65K Gemma Scope~2 SAEs; the corresponding repositories are listed in Table~\ref{tab:hf-repos-additional}.
We follow the same sampling and evaluation protocol described in Section~\ref{subsec:exp-config}.
Table~\ref{tab:additional-sae} reports Input and Output scores for all configurations.

\begin{table*}[ht]
  \caption{Input and Output scores (\%) across methods on five additional SAE configurations. Values are the mean score over layers along with their respective 95\% confidence intervals. For each (configuration, score-type) the largest value is in \textbf{bold}.}
  \label{tab:additional-sae}
  \begin{center}
    \begin{footnotesize}
      \setlength{\tabcolsep}{3pt}
      \begin{tabular}{l
        >{\columncolor{inputbg}}r r
        >{\columncolor{inputbg}}r r
        >{\columncolor{inputbg}}r r
        >{\columncolor{inputbg}}r r
        >{\columncolor{inputbg}}r r}
        \toprule
        & \multicolumn{2}{c}{\textbf{GPT-2 Small (128K)}} & \multicolumn{2}{c}{\textbf{Gemma-3-270M (16K)}} & \multicolumn{2}{c}{\textbf{Gemma-3-1B (16K)}} & \multicolumn{2}{c}{\textbf{Gemma-3-4B (16K)}} & \multicolumn{2}{c}{\textbf{Gemma-3-4B (65K)}} \\
        \cmidrule(lr){2-3} \cmidrule(lr){4-5} \cmidrule(lr){6-7} \cmidrule(lr){8-9} \cmidrule(lr){10-11}
        & Input (\%) & Output (\%) & Input (\%) & Output (\%) & Input (\%) & Output (\%) & Input (\%) & Output (\%) & Input (\%) & Output (\%) \\
        \midrule
        \llk     & $\text{6.44}_{\pm \text{0.99}}$  & $\text{3.21}_{\pm \text{0.34}}$  & $\text{2.76}_{\pm \text{1.38}}$  & $\text{1.55}_{\pm \text{0.55}}$  & $\text{3.45}_{\pm \text{1.30}}$  & $\text{5.67}_{\pm \text{1.01}}$  & $\text{2.57}_{\pm \text{1.05}}$  & $\text{5.20}_{\pm \text{1.16}}$  & $\text{2.66}_{\pm \text{1.28}}$  & $\text{2.26}_{\pm \text{0.74}}$  \\
        \llv     & $\text{7.21}_{\pm \text{1.15}}$  & $\text{11.75}_{\pm \text{0.59}}$ & $\text{1.65}_{\pm \text{0.92}}$  & $\text{5.46}_{\pm \text{0.83}}$  & $\text{1.82}_{\pm \text{0.75}}$  & $\text{9.61}_{\pm \text{1.04}}$  & $\text{2.84}_{\pm \text{1.09}}$  & $\text{6.08}_{\pm \text{1.23}}$  & $\text{1.91}_{\pm \text{0.99}}$  & $\text{2.81}_{\pm \text{0.79}}$  \\
        \tl      & $\text{5.36}_{\pm \text{1.00}}$  & $\text{11.32}_{\pm \text{0.59}}$ & $\text{1.76}_{\pm \text{0.97}}$  & $\text{4.96}_{\pm \text{0.80}}$  & $\text{1.82}_{\pm \text{0.74}}$  & $\text{9.58}_{\pm \text{1.03}}$  & $\text{2.94}_{\pm \text{1.15}}$  & $\text{6.04}_{\pm \text{1.23}}$  & $\text{1.99}_{\pm \text{1.00}}$  & $\text{2.79}_{\pm \text{0.77}}$  \\
        \zo      & $\text{2.44}_{\pm \text{0.69}}$  & $\text{3.21}_{\pm \text{0.44}}$  & $\text{5.25}_{\pm \text{1.96}}$  & $\text{4.44}_{\pm \text{0.82}}$  & $\text{10.76}_{\pm \text{2.69}}$ & $\text{6.81}_{\pm \text{1.09}}$  & $\text{9.90}_{\pm \text{2.51}}$  & $\text{4.80}_{\pm \text{1.12}}$  & $\text{5.79}_{\pm \text{2.06}}$  & $\text{1.76}_{\pm \text{0.68}}$  \\
        \tc{1}   & $\text{21.19}_{\pm \text{1.91}}$ & $\text{13.17}_{\pm \text{0.61}}$ & $\text{7.55}_{\pm \text{2.32}}$  & $\text{7.11}_{\pm \text{0.91}}$  & $\text{11.75}_{\pm \text{2.67}}$ & $\text{7.22}_{\pm \text{1.05}}$  & $\text{6.42}_{\pm \text{2.12}}$  & $\text{2.97}_{\pm \text{0.85}}$  & $\text{6.04}_{\pm \text{2.13}}$  & $\text{2.04}_{\pm \text{0.61}}$  \\
        \tc{5}   & $\text{22.01}_{\pm \text{1.94}}$ & $\text{12.51}_{\pm \text{0.59}}$ & $\text{7.54}_{\pm \text{2.29}}$  & $\text{8.99}_{\pm \text{1.01}}$  & $\text{13.08}_{\pm \text{2.83}}$ & $\text{9.26}_{\pm \text{1.12}}$  & $\text{10.90}_{\pm \text{2.64}}$ & $\text{5.99}_{\pm \text{1.17}}$  & $\text{8.59}_{\pm \text{2.42}}$  & $\text{3.24}_{\pm \text{0.79}}$  \\
        \tc{10}  & $\text{22.14}_{\pm \text{1.93}}$ & $\text{11.89}_{\pm \text{0.58}}$ & $\text{7.66}_{\pm \text{2.30}}$  & $\text{9.36}_{\pm \text{1.03}}$  & $\text{13.28}_{\pm \text{2.86}}$ & $\text{9.60}_{\pm \text{1.11}}$  & $\text{11.28}_{\pm \text{2.68}}$ & $\text{6.37}_{\pm \text{1.23}}$  & $\text{9.31}_{\pm \text{2.51}}$  & $\text{3.20}_{\pm \text{0.76}}$  \\
        \midrule
        \qlk     & $\textbf{32.16}_{\pm \text{2.16}}$ & $\text{1.44}_{\pm \text{0.27}}$  & $\textbf{17.30}_{\pm \text{3.28}}$ & $\text{0.86}_{\pm \text{0.42}}$  & $\textbf{14.71}_{\pm \text{2.90}}$ & $\text{2.32}_{\pm \text{0.64}}$  & $\textbf{14.70}_{\pm \text{2.91}}$ & $\text{2.02}_{\pm \text{0.64}}$  & $\textbf{11.14}_{\pm \text{2.67}}$ & $\text{0.89}_{\pm \text{0.47}}$  \\
        \qlv     & $\text{17.73}_{\pm \text{1.78}}$ & $\textbf{14.43}_{\pm \text{0.67}}$ & $\text{8.54}_{\pm \text{2.41}}$  & $\textbf{10.26}_{\pm \text{1.04}}$ & $\text{13.39}_{\pm \text{2.85}}$ & $\textbf{10.87}_{\pm \text{1.15}}$ & $\text{11.12}_{\pm \text{2.64}}$ & $\textbf{6.99}_{\pm \text{1.26}}$  & $\text{9.54}_{\pm \text{2.52}}$  & $\textbf{3.53}_{\pm \text{0.81}}$  \\
        \bottomrule
      \end{tabular}
    \end{footnotesize}
  \end{center}
\end{table*}

The patterns observed in the main results (Table~\ref{tab:main}) generalize to these additional configurations: \qlk\ achieves the highest Input Score and \qlv\ the highest Output Score across SAE widths and model sizes, with the relative gain over Logit Lens preserved.

\subsection{Transcoders}
\label{app:additional-transcoder}

\paragraph{Background.}
A \emph{transcoder} learns a sparse dictionary directly for the MLP residual update \cite{dunefsky2024transcodersinterpretablellmfeature}.
Unlike an SAE, which reconstructs a residual stream vector, a transcoder is trained to map the MLP input $h_{\text{mid}}$ to its MLP residual update $R_{\text{M}}$:
\begin{equation*}
\widehat{R}_{\text{M}}
= f\!\left(h_{\text{mid}} W_{\text{enc}}^{\top}\right) W_{\text{dec}}.
\label{eq:transcoder}
\end{equation*}
Each \emph{key feature} $k_i^l$ (row of $W_{\text{enc}}$) reads from the pre-MLP residual $h_{\text{mid}}^l$, and each \emph{value feature} $v_i^l$ (row of $W_{\text{dec}}$) is written into the residual stream as part of the MLP output at layer $l$.

\paragraph{Stream Transition for Transcoders.}
Query Lens extends to transcoders with a small change in the indexing of the stream transition.
For the forward dynamics, the value feature $v_i^l$ is written into the residual stream as an MLP output, so the transition product begins after the layer-$l$ MLP block:
\[
\frac{\partial y}{\partial a_i^l}
=
U^{\top}
\left[\prod_{k=l+1}^{L} (I + J_{\text{M}}^{k})(I + J_{\text{A}}^{k})\right]
v_i^l.
\]
For the backward dynamics, the key feature reads from $h_{\text{mid}}^l$ rather than $h_{\text{post}}^l$, so the prefix transition product terminates after the layer-$l$ attention block:
\[
\frac{\partial a_i^l}{\partial x}
=
(k_i^l)^{\top}
\left[\bigl(I + J_{\text{A}}^{l}\bigr)\prod_{k=1}^{l-1}(I + J_{\text{M}}^{k})(I + J_{\text{A}}^{k})\right]
\widehat{E}.
\]
The remaining elements of the framework---feature vectors and readout---carry over unchanged.

\begin{table}[ht]
  \caption{Input and Output scores (\%) on transcoders for Qwen-3 at three scales. Values are the mean score over layers along with their respective 95\% confidence intervals. For each (configuration, score-type) the largest value is in \textbf{bold}.}
  \label{tab:additional-transcoder}
  \centering
  \begin{footnotesize}
    \setlength{\tabcolsep}{3pt}
    \begin{tabular}{l
      >{\columncolor{inputbg}}r r
      >{\columncolor{inputbg}}r r
      >{\columncolor{inputbg}}r r}
      \toprule
      & \multicolumn{2}{c}{\textbf{Qwen-3-0.6B}} & \multicolumn{2}{c}{\textbf{Qwen-3-1.7B}} & \multicolumn{2}{c}{\textbf{Qwen-3-4B}} \\
      \cmidrule(lr){2-3} \cmidrule(lr){4-5} \cmidrule(lr){6-7}
      & Input (\%) & Output (\%) & Input (\%) & Output (\%) & Input (\%) & Output (\%) \\
      \midrule
      \llk     & $\text{0.24}_{\pm \text{0.18}}$  & $\text{0.01}_{\pm \text{0.02}}$  & $\text{0.65}_{\pm \text{0.48}}$  & $\text{0.13}_{\pm \text{0.18}}$  & $\text{5.23}_{\pm \text{1.55}}$  & $\text{0.29}_{\pm \text{0.17}}$  \\
      \llv     & $\text{0.39}_{\pm \text{0.51}}$  & $\text{6.14}_{\pm \text{0.52}}$  & $\text{0.19}_{\pm \text{0.20}}$  & $\text{1.77}_{\pm \text{0.39}}$  & $\text{0.56}_{\pm \text{0.43}}$  & $\text{1.32}_{\pm \text{0.32}}$  \\
      \tl      & $\text{0.54}_{\pm \text{0.59}}$  & $\text{5.90}_{\pm \text{0.51}}$  & $\text{0.66}_{\pm \text{0.67}}$  & $\text{1.75}_{\pm \text{0.39}}$  & $\text{0.82}_{\pm \text{0.58}}$  & $\text{1.24}_{\pm \text{0.30}}$  \\
      \zo      & $\text{3.65}_{\pm \text{1.05}}$  & $\text{0.74}_{\pm \text{0.31}}$  & $\text{2.75}_{\pm \text{0.89}}$  & $\text{0.17}_{\pm \text{0.12}}$  & $\text{0.79}_{\pm \text{0.66}}$  & $\text{0.36}_{\pm \text{0.26}}$  \\
      \tc{1}   & $\text{1.59}_{\pm \text{1.13}}$  & $\text{7.78}_{\pm \text{0.60}}$  & $\text{2.81}_{\pm \text{1.31}}$  & $\text{3.12}_{\pm \text{0.54}}$  & $\text{1.65}_{\pm \text{0.89}}$  & $\text{7.36}_{\pm \text{0.81}}$  \\
      \tc{5}   & $\text{1.81}_{\pm \text{1.20}}$  & $\text{7.88}_{\pm \text{0.60}}$  & $\text{2.85}_{\pm \text{1.33}}$  & $\text{3.11}_{\pm \text{0.52}}$  & $\text{2.03}_{\pm \text{1.04}}$  & $\textbf{7.48}_{\pm \text{0.82}}$  \\
      \tc{10}  & $\text{1.85}_{\pm \text{1.22}}$  & $\text{7.90}_{\pm \text{0.60}}$  & $\text{2.75}_{\pm \text{1.30}}$  & $\text{2.91}_{\pm \text{0.50}}$  & $\text{2.69}_{\pm \text{1.21}}$  & $\text{7.16}_{\pm \text{0.80}}$  \\
      \midrule
      \qlk     & $\textbf{14.54}_{\pm \text{2.75}}$ & $\text{0.07}_{\pm \text{0.08}}$  & $\textbf{14.55}_{\pm \text{2.76}}$ & $\text{0.01}_{\pm \text{0.02}}$  & $\textbf{11.08}_{\pm \text{2.25}}$ & $\text{0.04}_{\pm \text{0.06}}$  \\
      \qlv     & $\text{1.40}_{\pm \text{1.09}}$  & $\textbf{8.06}_{\pm \text{0.69}}$  & $\text{1.51}_{\pm \text{0.91}}$  & $\textbf{4.58}_{\pm \text{0.67}}$  & $\text{1.36}_{\pm \text{0.81}}$  & $\text{6.93}_{\pm \text{0.79}}$  \\
      \bottomrule
    \end{tabular}
  \end{footnotesize}
\end{table}

\paragraph{Results.}
We evaluate Query Lens on transcoders trained for Qwen-3 at three scales (0.6B, 1.7B, 4B).
Table~\ref{tab:additional-transcoder} reports Input and Output scores following the same protocol as the main experiments.
Across all three transcoder scales, Query Lens preserves the qualitative behavior observed for SAEs: \qlk\ leads on the input side and \qlv\ on the output side.
The sole exception is the output side of Qwen-3-4B, where Token Change marginally surpasses \qlv\ within overlapping confidence intervals.
Overall, the framework extends naturally to sparse dictionaries trained on MLP outputs rather than residual stream vectors.

\section{The Fidelity--Variance Tradeoff in Stream Transition Modeling}
\label{app:A}

As established in the main text, the full stream transition from a value feature $v_i^{\,l}$ at layer $l$ to the output logits encompasses all possible residual paths through the intervening blocks. Formally, the Jacobian of the logits $y$ with respect to the feature activation $a_i^{\,l}$ is given by:

\begin{equation*}
\frac{\partial y}{\partial a_{i}^{\,l}} = U^{\top} \left[ \prod_{k=l+1}^{L} \bigl(I + J_{\text{M}}^{k}\bigr)\bigl(I + J_{\text{A}}^{k}\bigr) \right] v_{i}^{\,l},
\label{eq:dy-da-final}
\end{equation*}

where the product spans $L-l$ layers, each containing two residual modules (attention and MLP). Expanding this product yields $2^{2(L-l)}$ additive terms, one for each way of choosing, per module, between its local Jacobian and the identity path.

Modeling this product in full is one extreme of a spectrum of choices for how much of the stream transition to include. At the other extreme, the Logit Lens (LL) keeps only the identity term, discarding all indirect effects. An intermediate, \emph{first-order} choice retains the identity and first-order Jacobian terms but drops the higher-order module interactions:

\begin{equation*}
\frac{\partial y}{\partial a_{i}^{\,l}} \approx U^{\top} \left[ I + \sum_{k=l+1}^{L} \left( J_{\text{M}}^{k} + J_{\text{A}}^{k} \right) \right] v_{i}^{\,l}.
\label{eq:dy-da-firstorder}
\end{equation*}

\paragraph{Two Regimes of Indirect-Effect Mediation.}
As discussed in Section~\ref{subsection:results-quant}, the benefit of mediating indirect effects grows as a feature moves farther from its stream endpoint, where the longer transition accumulates larger indirect contributions.
We break this down layer by layer to ask how much of the stream transition is worth modeling.
We sample 100 features at each of four source layers of Gemma-3-1B (16K SAE) and compare the output scores of the three strategies (LL, first-order, and full) on a common set of features.
Table~\ref{tab:stream-fidelity} reports the mean output score per strategy and an oracle that picks the best strategy for each feature.

The comparison reveals two regimes separated by the feature's distance from the output endpoint.
\emph{Far from the endpoint} (layers 7 and 13), the transition is long and most of the feature's effect on the logits flows through indirect paths; modeling them is essential, and the first-order and full transitions score several times higher than LL.
\emph{Near the endpoint} (layer 22), the transition is short and the direct projection already dominates; here mediating indirect effects no longer pays off, LL is best, and first-order and then full are progressively worse.

\begin{table}[ht]
\centering
\caption{Stream transition fidelity on Gemma-3-1B (16K SAE): mean output score per strategy (LL / first-order (FO) / full) and the relative variance of the FO and full transitions, for 100 features sampled at each source layer. ``Oracle'' picks the best strategy per feature ($\Delta$: gap over the best static layer-wise strategy). Layers run from far (long transition) to near (short) the output endpoint.}
\label{tab:stream-fidelity}
\small
\setlength{\tabcolsep}{5pt}
\begin{tabular}{lcccccc}
\toprule
 & \multicolumn{4}{c}{Output score (mean)} & \multicolumn{2}{c}{Rel.\ variance ($\downarrow$)} \\
\cmidrule(lr){2-5} \cmidrule(lr){6-7}
Layer & LL & FO & Full & Oracle ($\Delta$) & FO & Full \\
\midrule
7 (far)   & 0.010 & \textbf{0.046} & \textbf{0.046} & 0.059 ($+0.013$) & \textbf{0.32} & 0.89 \\
13        & 0.038 & \textbf{0.072} & \textbf{0.072} & 0.093 ($+0.021$) & \textbf{0.33} & 0.71 \\
17        & 0.154 & \textbf{0.174} & 0.161 & 0.206 ($+0.032$) & \textbf{0.24} & 0.51 \\
22 (near) & \textbf{0.183} & 0.176 & 0.156 & 0.213 ($+0.030$) & \textbf{0.13} & 0.20 \\
\bottomrule
\end{tabular}
\end{table}

\paragraph{A Variance Cost of Indirect-Effect Mediation.}
We attribute the second regime to the \emph{variance} that indirect-effect mediation injects into the linearization.
Each transition maps a feature to a vector $g$ that depends on the sampled hidden state; treating the token-wise vectors $\{g_n\}$ as samples, we measure their context-sensitivity with the Relative Variance
\begin{equation*}
\text{Relative Variance} = \frac{\mathbb{E}\!\left[\|g - \mu\|^{2}\right]}{\mathbb{E}\!\left[\|g\|^{2}\right]}, \quad \mu = \mathbb{E}[g],
\label{eq:relative-variance}
\end{equation*}
where a value near $0$ means the transition is essentially context-invariant and a value near $1$ means it varies strongly across contexts.
The last two columns of Table~\ref{tab:stream-fidelity} show that the full transition has much higher relative variance than the first-order one at every layer (e.g., $0.89$ vs.\ $0.32$ at layer 7), and that this variance grows sharply with transition length, from $0.20$ near the endpoint to $0.89$ far from it; LL, a fixed projection, carries no transition variance at all.

This variance cost can be derived analytically.
For i.i.d.\ scalar Jacobians $J^k$ with mean $\mu$ and variance $\sigma^2$ composed across $L$ modules, the full and first-order transitions have variance ratio
\begin{equation*}
\frac{\operatorname{Var}\!\left[\prod_{k=1}^{L}\bigl(1+J^k\bigr)\right]}{\operatorname{Var}\!\left[1+\sum_{k=1}^{L}J^k\right]}
= \frac{(\alpha+\sigma^2)^{L} - \alpha^{L}}{L\,\sigma^2}, \qquad \alpha=(1+\mu)^2,
\label{eq:var-ratio}
\end{equation*}
which is strictly greater than $1$ for all $L\geq 2$ and $\sigma^2>0$ and grows monotonically in $L$: every higher-order cross-term adds variance with no counterpart in the first-order sum.
The full transition is therefore inherently noisier than its first-order truncation, increasingly so for longer transitions.

Together this gives a signal-versus-variance reading: mediating indirect effects helps only when their signal outweighs the variance it adds. This holds far from the endpoint but not near it, where the signal is negligible and the variance-free LL wins.

\paragraph{First-Order Mediation as an Intermediate Strategy.}
The first-order transition is a fair middle ground: it captures the dominant first-order indirect signal at roughly a third of the full transition's relative variance.
As a result it matches or exceeds the full transition on output score at every layer in Table~\ref{tab:stream-fidelity} (it is the single best strategy at layer 17), while never trailing LL by much.
It is not, however, uniformly optimal: near the endpoint LL is still better, and on individual features the full transition remains preferable.
First-order mediation is thus a reasonable static default rather than a universal answer.

\paragraph{Toward Adaptive Fidelity.}
\label{app:layer-analysis}
The oracle quantifies the headroom left by any static choice: picking the best strategy \emph{per feature} adds a further $+0.013$ to $+0.032$ over the best static layer-wise strategy at every layer, and the best static strategy itself flips with depth.
No single fidelity is right for all features.
This calls for an adaptive strategy that decides, per feature, how much of the stream transition to mediate: a faithful low-variance approximation where the indirect signal is weak, and the full transition where it is strong.
We leave the design of such per-feature fidelity selection to future work.

\section{Pre-Activation as a Proxy for Post-Activation}
\label{app:pre-act-proxy}

The backward dynamics of Section~3.2.2 measure a feature's input-side sensitivity by differentiating its \emph{pre-activation} with respect to the input.
All SAEs in our study (Section~4.1) rely on either JumpReLU or TopK activations, both of which involve discontinuous thresholding operations and thus yield gradients ill-suited to this purpose.
We show that differentiating the pre-activation instead recovers a well-defined input direction that reflects which input perturbations would activate the feature.

\paragraph{Pre-Activation Gives a Well-Defined Input Direction.}
Both nonlinearities apply a threshold that zeros out insufficiently active features.
Gradients of zeroed-out features are zero; therefore, remain uninformative about which input perturbations would bring the feature to life.
However, the pre-activation gradient does not suffer this collapse: it remains informative even when the feature is not largely active, and also indicates the input perturbations that would push it toward activation.
Moreover, increasing the pre-activation is exactly what is required to increase the post-activation, because the activation function is merely a gate, without altering the underlying relationship between the input and the pre-activation.

\paragraph{Connection to Straight-Through Estimators.}
Differentiating the pre-activation rather than the post-activation can be viewed as applying a straight-through estimator \cite{bengio2013estimating}: the forward pass respects the discontinuous activation, while the backward pass treats it as the identity and lets gradients flow through unimpeded.
Replacing this hard estimator with a smooth continuous relaxation, such as Gumbel-Softmax \cite{jang2017categorical}, is a promising direction for future work.

\section{Efficient Computation of Query Lens}
\label{app:ql-compute}

The Query Lens scores of Eq.~\eqref{eq:s_value} and its key counterpart both apply the stream transition $\prod_k (I + J^{k})$, a $d_{\mathrm{m}} \times d_{\mathrm{m}}$ map, to a feature vector. Read at face value, the definition suggests two sources of cost: building the transition Jacobian, and summing the many terms it produces when the product is expanded. We incur neither. Each score reduces to a single forward or backward pass under automatic differentiation, at the cost of one ordinary pass through the relevant layers.

\paragraph{Jacobian Products.}
The transition never stands on its own. The value score left-multiplies it onto $v_i^{\,l}$ and the key score right-multiplies $(k_i^{\,l})^{\top}$ onto it, so in both cases we only need the transition applied to one vector. This is precisely a Jacobian--vector product (JVP, on the value side) or a vector--Jacobian product (VJP, on the key side), which automatic differentiation returns in a single pass without instantiating the Jacobian. Forming the full $d_{\mathrm{m}} \times d_{\mathrm{m}}$ transition would instead take $d_{\mathrm{m}}$ such passes, one per column, and $O(d_{\mathrm{m}}^{2})$ memory to hold it; the JVP and VJP need one pass and $O(d_{\mathrm{m}})$ memory for the running vector.

\paragraph{In-Place Accumulation.}
Expanding $\prod_k (I + J^{k})$ yields a separate term for every computational path through the residual stream, a count exponential in the number of blocks, which might suggest computing and summing these path terms one by one. We never form this sum. The JVP carries a single tangent and the VJP a single cotangent, updated one block at a time, $u \leftarrow u + J^{k} u$ on the value side and $w \leftarrow w + (J^{k})^{\top} w$ on the key side. Each update applies one factor $(I + J^{k})$ to the running vector, so the contribution of every path is accumulated in place as the vector is transported across layers. The transport therefore costs $O(L)$ factor applications rather than the $O(2^{L})$ terms of the explicit expansion.

\section{Robustness Analysis}
\label{app:B}

\subsection{Data Invariance of Predicted Token Sets}
\label{app:B2}

Because the stream transition in Query Lens is computed from hidden states of a sampled corpus (Appendix~\ref{app:baseline-impl}), we test whether the predicted token sets change with the choice of sample.
We draw three independent Pile subsets with different random seeds ($42$, $17$, $123$), run Query Lens on each to obtain top-$k$ token sets $T_1, T_2, T_3$, and measure their overlap with the three-way Jaccard similarity
\begin{equation*}
J(T_1,T_2,T_3) = \frac{\left|T_1 \cap T_2 \cap T_3\right|}{\left|T_1 \cup T_2 \cup T_3\right|}.
\label{eq:jaccard-3way}
\end{equation*}

\begin{table}[ht]
  \caption{Data invariance of predicted token sets for GPT-2 Small at two SAE widths, measured by the three-way Jaccard similarity $J(T_1,T_2,T_3)$ across three independently sampled Pile subsets (mean $\pm$ standard error across features).}
  \label{tab:data-invariance-jaccard}
  \centering
  \begin{small}
    \begin{tabular}{lcc}
      \toprule
      SAE width & \qlk & \qlv \\
      \midrule
      32K  & $\text{0.8492}_\text{$\pm 0.0027$}$ & $\text{0.8879}_\text{$\pm 0.0027$}$ \\
      128K & $\text{0.8317}_\text{$\pm 0.0028$}$ & $\text{0.8845}_\text{$\pm 0.0026$}$ \\
      \bottomrule
    \end{tabular}
  \end{small}
  \vskip -0.1in
\end{table}

Table~\ref{tab:data-invariance-jaccard} reports this for GPT-2 Small at two SAE widths.
Both variants exceed $0.83$ at both SAE widths: the token sets recovered from independent corpora are largely shared, indicating that a Query Lens explanation is a property of the feature itself rather than of the particular sample used to compute it.
This is what sets Query Lens apart from activation-based methods, which instead read a feature's meaning off the content of selected input examples.

\subsection{Sensitivity to the Number of Top Tokens $k$}
\label{app:k-sensitivity}
Throughout the main experiments we fix the explanation size to the top $k=25$ tokens (Section~\ref{subsec:exp-config}).
We verify that this choice does not drive our conclusions.
For each of the four configurations in Table~\ref{tab:main}, we recompute the Input and Output scores while varying $k\in\{5,10,20,25\}$ under the identical metric, comparing QL against the matched LL baseline (the key variant for the Input score and the value variant for the Output score).

Table~\ref{tab:k-sensitivity} reports the results.
Two patterns hold across all settings.
First, both scores grow monotonically with $k$ for every method, since a larger token set $T$ is more likely to cover the target set.
Second, the ordering between QL and LL holds at every $k$: QL leads on the input side by a large margin (often by an order of magnitude), and on the output side at $k=25$ and almost all smaller $k$. The only exception is a marginal reversal for Qwen-3-1.7B at $k=5$ ($6.50$ vs.\ $6.65$).
QL's gains over LL therefore do not hinge on the specific choice of $k=25$.

\begin{table*}[ht]
  \caption{Sensitivity of Input and Output scores (\%) to the number of top tokens $k$, for the four configurations of Table~\ref{tab:main}. Input scores use the key variant (\llk, \qlk) and Output scores the value variant (\llv, \qlv). For each (configuration, score, $k$) the larger of QL and LL is in \textbf{bold}.}
  \label{tab:k-sensitivity}
  \centering
  \begin{footnotesize}
    \begin{tabular}{lc cccc cccc}
      \toprule
      & & \multicolumn{4}{c}{\textbf{Input Score (\%)}} & \multicolumn{4}{c}{\textbf{Output Score (\%)}} \\
      \cmidrule(lr){3-6} \cmidrule(lr){7-10}
      Configuration & Method & $k{=}5$ & $k{=}10$ & $k{=}20$ & $k{=}25$ & $k{=}5$ & $k{=}10$ & $k{=}20$ & $k{=}25$ \\
      \midrule
      \multirow{2}{*}{GPT-2 Small (32K)} & QL & \textbf{30.37} & \textbf{34.73} & \textbf{38.12} & \textbf{39.32} & \textbf{12.52} & \textbf{13.62} & \textbf{14.97} & \textbf{15.24} \\
       & LL & 4.50 & 5.49 & 7.33 & 7.84 & 10.63 & 11.45 & 12.50 & 12.57 \\
      \midrule
      \multirow{2}{*}{Gemma-3-270M (65K)} & QL & \textbf{16.75} & \textbf{19.25} & \textbf{20.81} & \textbf{21.82} & \textbf{5.35} & \textbf{7.98} & \textbf{10.16} & \textbf{10.63} \\
       & LL & 0.15 & 0.45 & 0.47 & 0.71 & 3.25 & 4.83 & 5.81 & 6.13 \\
      \midrule
      \multirow{2}{*}{Gemma-3-1B (65K)} & QL & \textbf{10.81} & \textbf{12.45} & \textbf{13.46} & \textbf{14.14} & \textbf{5.90} & \textbf{7.10} & \textbf{8.84} & \textbf{9.26} \\
       & LL & 1.09 & 1.35 & 1.66 & 1.74 & 5.05 & 6.05 & 7.45 & 7.84 \\
      \midrule
      \multirow{2}{*}{Qwen-3-1.7B (32K)} & QL & \textbf{14.75} & \textbf{17.90} & \textbf{20.72} & \textbf{21.69} & 6.50 & \textbf{7.78} & \textbf{9.28} & \textbf{9.36} \\
       & LL & 0.85 & 1.38 & 1.66 & 1.91 & \textbf{6.65} & 7.27 & 8.16 & 8.31 \\
      \bottomrule
    \end{tabular}
  \end{footnotesize}
\end{table*}

\section{Evaluation Details}
\label{app:eval-details}

\subsection{Activation Data}
\label{app:activation-data}

For each feature, the set of sentences $S$ used in the input-side evaluation (Section~\ref{subsection:input-side}) is constructed as follows.
For GPT-2 Small and Gemma-3 features, we use the pre-computed top-activating examples from Neuronpedia~\cite{neuronpedia}.
For Qwen-3-1.7B features, Qwen-Scope SAEs are not indexed by Neuronpedia, so we collect activations manually: we run the model on $16384\times 128$ tokens randomly sampled from \texttt{monology/pile-uncopyrighted}\footnote{\url{https://huggingface.co/datasets/monology/pile-uncopyrighted}} and retain the top-$20$ activating contexts per feature.

\subsection{Steering Details}
\label{app:C}

\paragraph{Choice of steering strengths $\mathcal{A}$.}
For each feature, we select the steering strength $\alpha$ to target a fixed KL divergence between the clean and steered next-token distributions \cite{gur-arieh-etal-2025-enhancing, paulo2025automaticallyinterpretingmillionsfeatures}, yielding $\mathcal{A}=\{\alpha_{\mathrm{KL}=0.25},\,\alpha_{\mathrm{KL}=0.5}\}$. KL targeting normalises the steering intensity across features that have very different natural activation scales.

\paragraph{Generation setup.}
Following \citet{arad-etal-2025-saes}, we generate 20 tokens using a temperature of 0.75 after each of the prefixes listed in Table~\ref{tab:neutral_prefixes}.  
While the original study uses 50 prefixes, we augment the set to 100 prefixes generated with GPT-5.2 \cite{2025gpt5} to increase prompt diversity.  

\begin{table}[ht]
\caption{Neutral prefixes used for generation in the steering experiments.}
\label{tab:neutral_prefixes}
\centering
\small
\setlength{\tabcolsep}{6pt}
\begin{tabularx}{\linewidth}{|X|X|X|}
\hline &&
\\
``Findings show that'' &
``It's no surprise that'' &
``It's been a long time since'' \\

``I once heard that'' &
``Have you ever noticed that'' &
``In my experience,'' \\

``Then the man said:" &
``I couldn't believe when" &
``The craziest part was when" \\

``I believe that" &
``The first thing I heard was" &
``If you think about it," \\

``The news mentioned" &
``Let me tell you a story about" &
``I was shocked to learn that" \\

``She saw a" &
``Someone once told me that" &
``For some reason," \\

``It is observed that" &
``It might sound strange, but" &
``I can't help but wonder if" \\

``Studies indicate that" &
``They always warned me that" &
``It makes sense that" \\

``According to reports," &
``Nobody expected that" &
``At first, I didn't believe that" \\

``Research suggests that" &
``Funny thing is," &
``That reminds me of the time when" \\

``It has been noted that" &
``I never thought I'd say this, but" &
``It all comes down to" \\

``I remember when" &
``What surprised me most was" &
``One time, I saw that" \\

``It all started when" &
``The other day, I overheard that" &
``I was just thinking about how" \\

``The legend goes that" &
``Back in the day," &
``Imagine a world where" \\

``If I recall correctly," &
``You won't believe what happened when" &
``They never expected that" \\

``People often say that" &
``A friend of mine once said," &
``I always knew that" \\

``Once upon a time," &
``I just found out that" &
``Over the years, I noticed that" \\

``Looking back, I realize that" &
``From what I can tell," &
``As far as I know," \\

``For a long time, I thought that" &
``It all made sense when" &
``The strange thing is that" \\

``People rarely talk about how" &
``I recently discovered that" &
``To my surprise," \\

``Every now and then, I notice that" &
``There is growing evidence that" &
``The more I learn, the more I see that" \\

``Experts often point out that" &
``I still remember the moment when" &
``I used to believe that" \\

``Over time, it became clear that" &
``It started to make sense when" &
``Many people forget that" \\

``What nobody told me was that" &
``You might have noticed that" &
``It was only later that I realized" \\

``It felt like" &
``I keep coming back to the idea that" &
``For as long as I can remember," \\

``It suddenly hit me that" &
``The more I think about it," &
``I have a feeling that" \\

``Some people argue that" &
``There is a common belief that" &
``Lately, I have been noticing that" \\

``One pattern I keep seeing is that" &
``Some say that" &
``Over time, people began to realize that" \\

``I started to notice that" &
``From time to time, I hear that" &
``It became obvious that" \\

``There was a moment when I realized that" &
``It turned out that" &
``What's interesting is that" \\

``Looking around, you can see that" &
``You could say that" &
``It often happens that" \\

``For many people, it seems that" &
``Little did I know that" & 
``It often turns out that" \\

``I get the sense that" &
``One thing that stands out is that" &
``If history has taught us anything, it is that" \\

``Something I have been wondering about is" &
 & \\ & &
\\
\hline
\end{tabularx}
\end{table}

\section{Additional Details on Subspace Channel Experiments}
\label{app:D}
\subsection{Training Details for Low-Rank Mappings}
\label{app:D.1}
This section provides training details for the low-rank linear maps $W^{l \rightarrow k}$ used to model feature reading channels in Section~\ref{sec:subspace-channel}.

\paragraph{Data and Regression Setup.}
For each source layer $l \in \{1, \ldots, L\}$ of GPT-2 Small ($L=12$), we randomly sample $N = 1000$ value features $\{v_i^l\}_{i=1}^{1000}$ from the GPT-2 Small SAE 32k. For each sampled feature $v_i^l$, we compute its downstream module responses
\begin{equation*}
    r_i^{l\rightarrow k} = J^k v_i^l \in \mathbb{R}^{d_m},
\end{equation*}
for all destinations $k > l$, where $J^k$ denotes the Jacobian of the residual block at layer $k$ with respect to the residual stream. This yields $(v_i^l, r_i^{l\rightarrow k})$ for each ordered pair $(l, k)$, which defines a supervised regression problem from value features to module responses.

\paragraph{Training.}
For each pair $(l, k)$ we split the samples independently into a training set (90\%) and a validation set (10\%). As described in Section~\ref{sec:subspace-channel}, we then train a low-rank matrix $W^{l\rightarrow k} \in \mathbb{R}^{d_m \times d_m}$ of rank $r=d_m/L=64$ to minimize the reconstruction error between predicted and true module responses:
\begin{equation*}
    \mathcal{L}_{\text{MSE}}=\left\Vert r_i^{l\rightarrow k} - W^{l\rightarrow k}v_i^l\right\Vert_2^2 .
\end{equation*}
Training is performed independently for each pair $(l, k)$ with the Adam optimizer (learning rate $\eta = 1\times 10^{-3}$, no weight decay) for 1000 epochs at \texttt{bfloat16} precision, using a single gradient-accumulation step and random seed 42.

\paragraph{Evaluation.}
We evaluate each map with four metrics: mean squared error (MSE, $\left\Vert r - \hat{r}\right\Vert_2^2$), relative L2 error ($\left\Vert r - \hat{r}\right\Vert_2/\left\Vert r\right\Vert_2$), cosine similarity ($r\cdot \hat{r} / \left\Vert r\right\Vert_2\left\Vert \hat{r}\right\Vert_2$), and explained energy ratio ($1 - \left\Vert r - \hat{r}\right\Vert_2^2/\left\Vert r\right\Vert_2^2$). Table~\ref{tab:recon-metrics} reports these averaged over the $66$ learned maps. The reconstruction generalizes well to held-out features: on the validation split it attains high cosine similarity and explains over half of the response energy despite the severe rank reduction.

\begin{table}[ht]
\centering
\caption{Reconstruction performance averaged over 66 mappings.}
\label{tab:recon-metrics}
\begin{tabular}{lcccc}
\toprule
\textbf{Split} & \textbf{MSE} & \textbf{Rel.\ L2 Error} & \textbf{Cosine Sim.} & \textbf{Explained Energy} \\
\midrule
Train & 0.000094 & 0.5540 & 0.7987 & 0.6777 \\
Val   & 0.000152 & 0.6500 & 0.7273 & 0.5691 \\
\bottomrule
\end{tabular}
\end{table}

\subsection{Statistical Tests for $\mathrm{OL}$}
\label{app:D.2}
Figure~\ref{fig:overlap-distribution} suggests that overlaps between maps sharing the same destination layer are systematically higher than overlaps between maps with different destinations. We quantify this separation with a permutation test that keeps the full set of learned maps fixed while randomizing the grouping.

\paragraph{Grouping and Observed Gap.}
The $66$ learned maps across $12$ layers form $2145$ pairs $(W^{l_1\rightarrow k_1},W^{l_2\rightarrow k_2})$, of which $220$ share the same destination layer ($k_1=k_2$) and $1925$ have different destinations ($k_1\ne k_2$). Table~\ref{tab:overlap-stats} reports the mean and standard deviation of each group, an observed mean gap of $\Delta_\text{mean}=0.4161$.

\begin{table}[ht]
    \centering
    \caption{Value of mean and standard deviation of $\mathrm{OL}(W_i, W_j)$ from pairs in each group.}
    \begin{tabular}{l c c}
        \toprule
        \textbf{Group} & \textbf{Mean} & \textbf{Std.} \\
        \midrule
        $k_1=k_2$ & 0.5670 & 0.0516  \\
        $k_1\ne k_2$ & 0.1509 & 0.0480 \\
        \bottomrule
    \end{tabular}
    \label{tab:overlap-stats}
\end{table}

\paragraph{Permutation Test.}
To assess significance without parametric assumptions, we run a one-sided permutation test: we keep the learned maps fixed and randomly shuffle the destination-layer labels $k$, redefining the two groups on every permutation. For each shuffled assignment, we recompute (1) the mean gap $\Delta_\text{mean}$ and (2) separation statistic AUC. The $p$-value is estimated as the fraction of permutations whose statistic is at least as large as the observed value. With $N_\text{perm}=10{,}000$, we obtain $p=0.0001$ for both $\Delta_\text{mean}$ and AUC, indicating that the observed separation is highly unlikely under random destination assignments.



\section{Qualitative Analysis}
\label{app:G}
\subsection{Prompt Details}
\label{app:G1}

\refstepcounter{table}\label{tab:prompt}%
\noindent
\begin{tcolorbox}[
    breakable,
    enhanced jigsaw,
    title={Prompt for Semantic Summarization and Interpretability Scoring},
    colback=white,
    colframe=black!25,
    fonttitle=\bfseries,
    boxrule=0.5pt,
    coltitle=black,
    sharp corners,
    arc=0pt,
    left=4pt, right=4pt, top=6pt, bottom=6pt
]
You will be given a ranked list of vocabulary tokens that are most strongly promoted in the output (or most strongly activating on the input) of a single sparse feature in a language model. Your job is to judge how interpretable this feature is.\par
\par\vspace{0.4em}
Do two things:\par
1) \textbf{Meaning summary}: name the single dominant theme that the tokens support.\par
2) \textbf{Interpretability score} (0--10): rate the \emph{topical concentration} of the token list --- i.e., what fraction of tokens directly support one coherent theme.\par
\par\vspace{0.6em}

\textbf{How to read the tokens}\par
- Tokens may be subword/BPE pieces (e.g., ``ansas'' from ``Kansas'', `` Direct'' from ``Director''). Consider the canonical word or concept the piece belongs to before judging coherence.\par
- Multiple languages or scripts referring to the same concept (e.g., ``food'' in English, Chinese, Thai, Korean) count as ONE coherent theme. Do not penalize multilingual coverage; treat it the same as monolingual coverage of the same concept.\par
- Tokens are typically rank-ordered by promotion strength. Earlier tokens carry slightly more weight, but treat the full list when computing concentration.\par
- If two coherent themes coexist, pick the dominant one for Summary, but do not lower the score solely because more than one theme is present --- judge on coverage of the dominant theme.\par
\par\vspace{0.6em}

\textbf{Scoring rubric} (use the \% of tokens directly supporting the dominant theme)\par
- 9--10: $\geq$80\% of tokens directly support a single, specific theme. Off-topic tokens are rare and look like residual noise.\par
- 7--8: 60--79\% support; theme is clear; noticeable but minority off-topic tokens.\par
- 4--6: 40--59\% support; theme is identifiable but $\sim$half the list is mixed or off-topic.\par
- 2--3: 20--39\% support; weak clustering; theme is plausible but much of the list is unrelated.\par
- 0--1: $<$20\% support, or no recognizable specialized theme. Tokens look generic, scattered, or like punctuation/format noise.\par
\par\vspace{0.6em}

\textbf{Calibration examples} (do not echo these in your output)\par
\emph{Example A --- Score 9.}\par
Tokens: `` meal,  meals,  food,  foods,  cuisine,  gastronomy, $\langle$zh:meal$\rangle$, $\langle$th:food$\rangle$, $\langle$ko:food$\rangle$,  comida,  Mahlzeit,  cibo,  repas,  makanan,  yemek,  meal time,  dining,  edible,  nourishment,  appetite,  snack,  feast,  buffet,  culinary,  diet''\par
Summary: ``Food and meals across multiple languages.''\par
Explanation: ``Almost every token names food, a meal, or eating; the multilingual variants describe the same concept.''\par
$\rightarrow$ 9\par
\par\vspace{0.4em}
\emph{Example B --- Score 6.}\par
Tokens: `` manager,  Director,  supervisor,  Manager,  director,  CEO,  chief,  Co,  ord, inate,  Sup, ervis, or,  staff,  team,  office,  colleague,  the,  and,  with,  to,  of,  in,  on, ed''\par
Summary: ``Senior management / leadership roles.''\par
Explanation: ``About half the tokens denote leadership/management roles; the remainder are common function words and BPE fragments unrelated to the theme.''\par
$\rightarrow$ 6\par
\par\vspace{0.4em}
\emph{Example C --- Score 1.}\par
Tokens: a scattered mix of unrelated fragments across Arabic, CJK, Devanagari, Sinhala, Tamil, Bengali, Kannada, Cyrillic, and Vietnamese scripts together with German BPE pieces and punctuation/format symbols.\par
Summary: ``None''\par
Explanation: ``Tokens are scattered across unrelated languages, punctuation, and format pieces with no recognizable single theme.''\par
$\rightarrow$ 1\par
\par\vspace{0.6em}

\textbf{Rules}\par
- Be strict. Topical concentration, not confidence in your guess, drives the score.\par
- If no specific theme is supported, set Summary to ``None'' and give a score in 0--1.\par
- Do not infer intent, sentiment, or specific entity identities beyond what the tokens directly support.\par
\par\vspace{0.6em}

\textbf{Output}\par
Return RAW JSON only. No markdown fences, no commentary outside the JSON.\par
\par\vspace{0.4em}
\{ \par
\ \ ``Score'': $<$integer 0 to 10$>$,\par
\ \ ``Summary'': ``$<$one-sentence specific meaning/function or `None'$>$'',\par
\ \ ``Explanation'': ``$<$one short sentence: which tokens support the theme and roughly what fraction of the list does$>$''\par
\}
\end{tcolorbox}

\subsection{Qualitative Examples}
\label{app:G2}
\begin{figure*}[p]
  \centering
  \includegraphics[width=\linewidth]{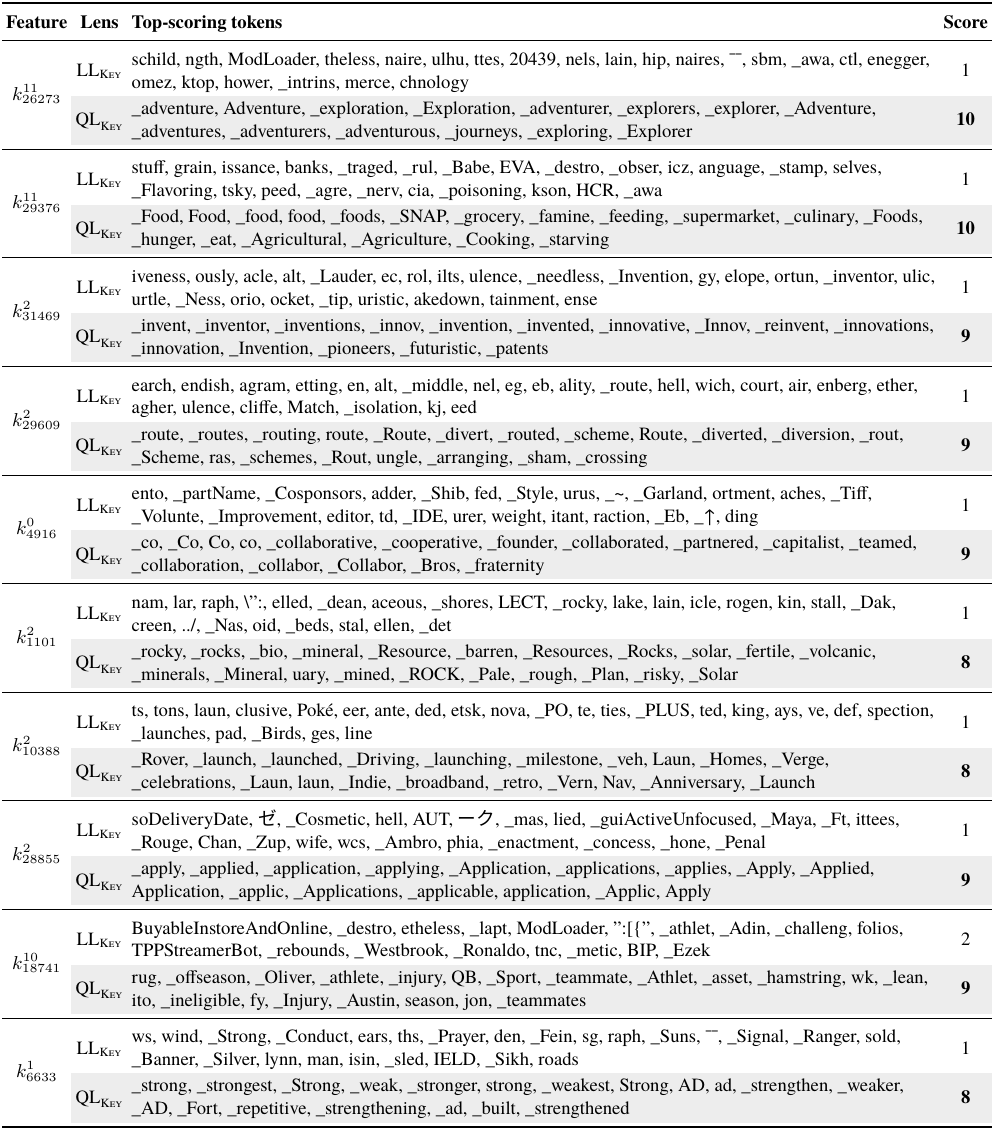}
  \caption{Qualitative \textbf{key} feature examples on GPT-2 Small (32K).}
  \label{fig:qualJ-gpt2-key}
\end{figure*}

\begin{figure*}[p]
  \centering
  \includegraphics[width=\linewidth]{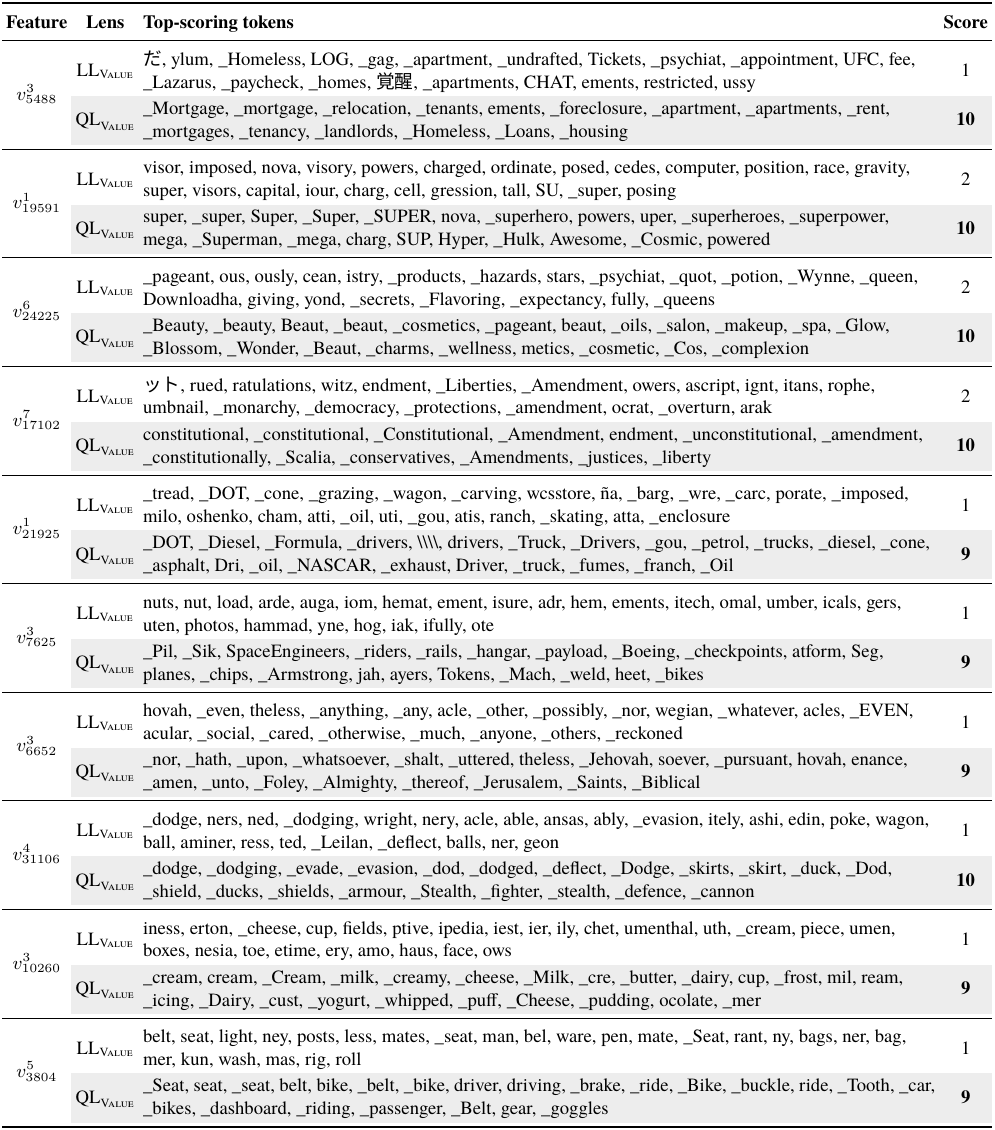}
  \caption{Qualitative \textbf{value} feature examples on GPT-2 Small (32K).}
  \label{fig:qualJ-gpt2-value}
\end{figure*}

\begin{figure*}[p]
  \centering
  \includegraphics[width=\linewidth]{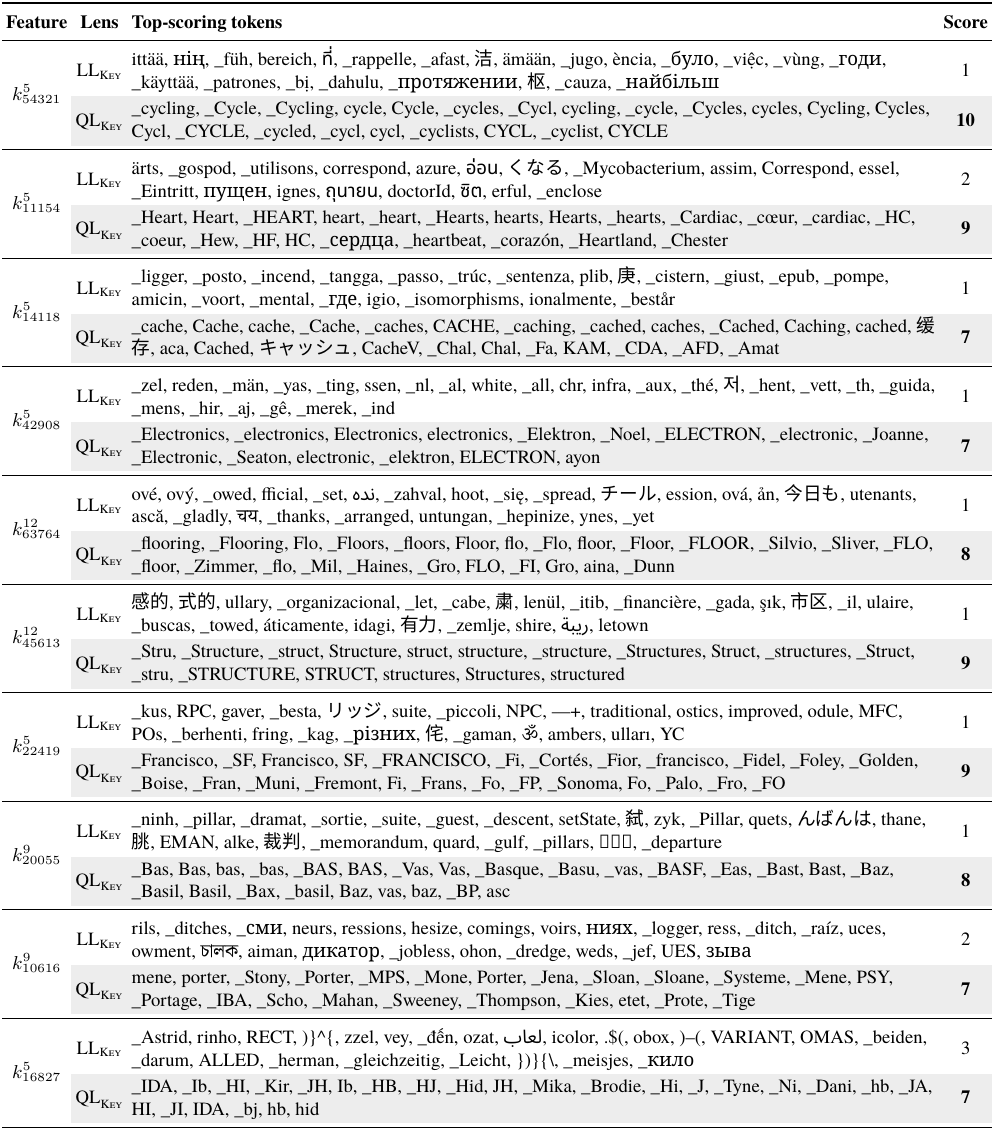}
  \caption{Qualitative \textbf{key} feature examples on Gemma-3-270M (65K).}
  \label{fig:qualJ-g270m-key}
\end{figure*}

\begin{figure*}[p]
  \centering
  \includegraphics[width=\linewidth]{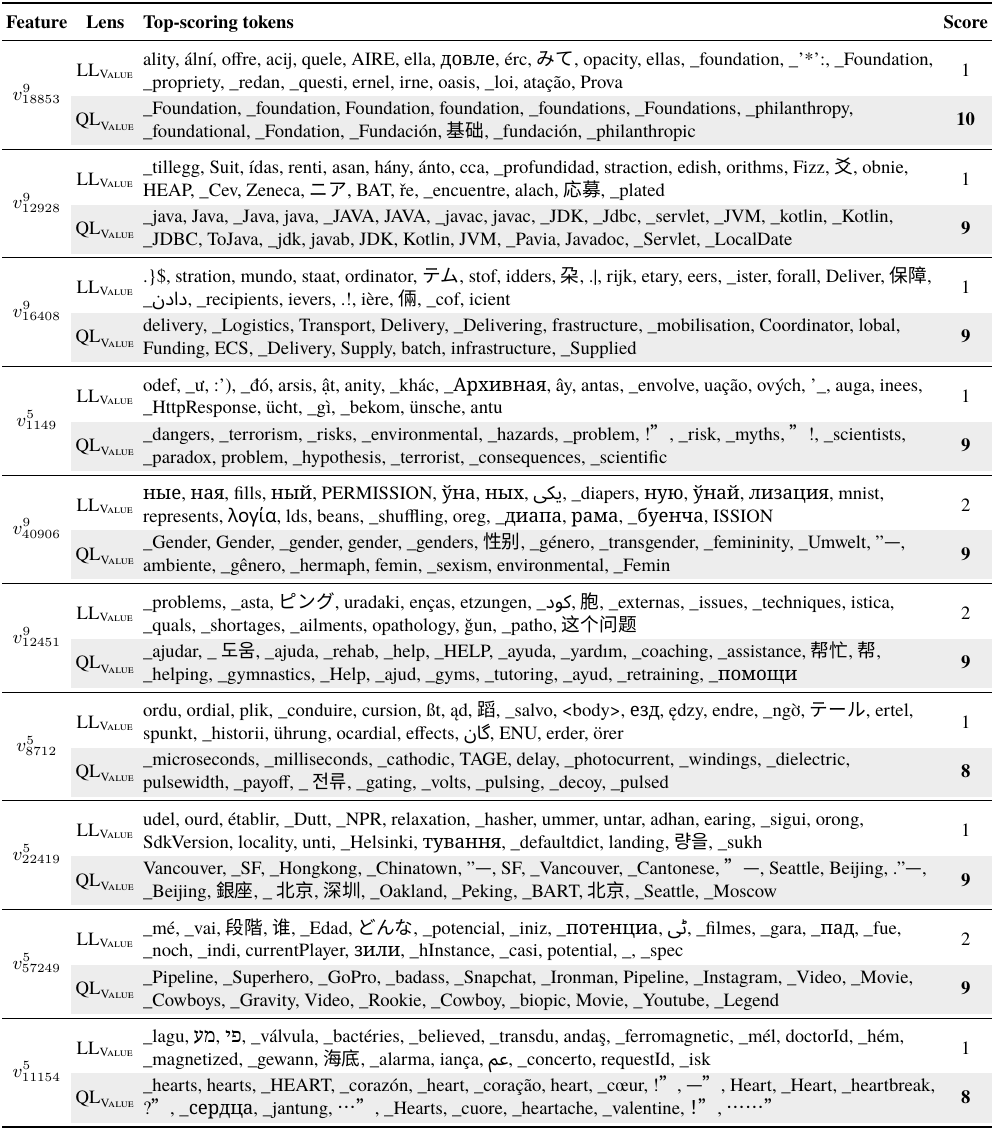}
  \caption{Qualitative \textbf{value} feature examples on Gemma-3-270M (65K).}
  \label{fig:qualJ-g270m-value}
\end{figure*}

\begin{figure*}[p]
  \centering
  \includegraphics[width=\linewidth]{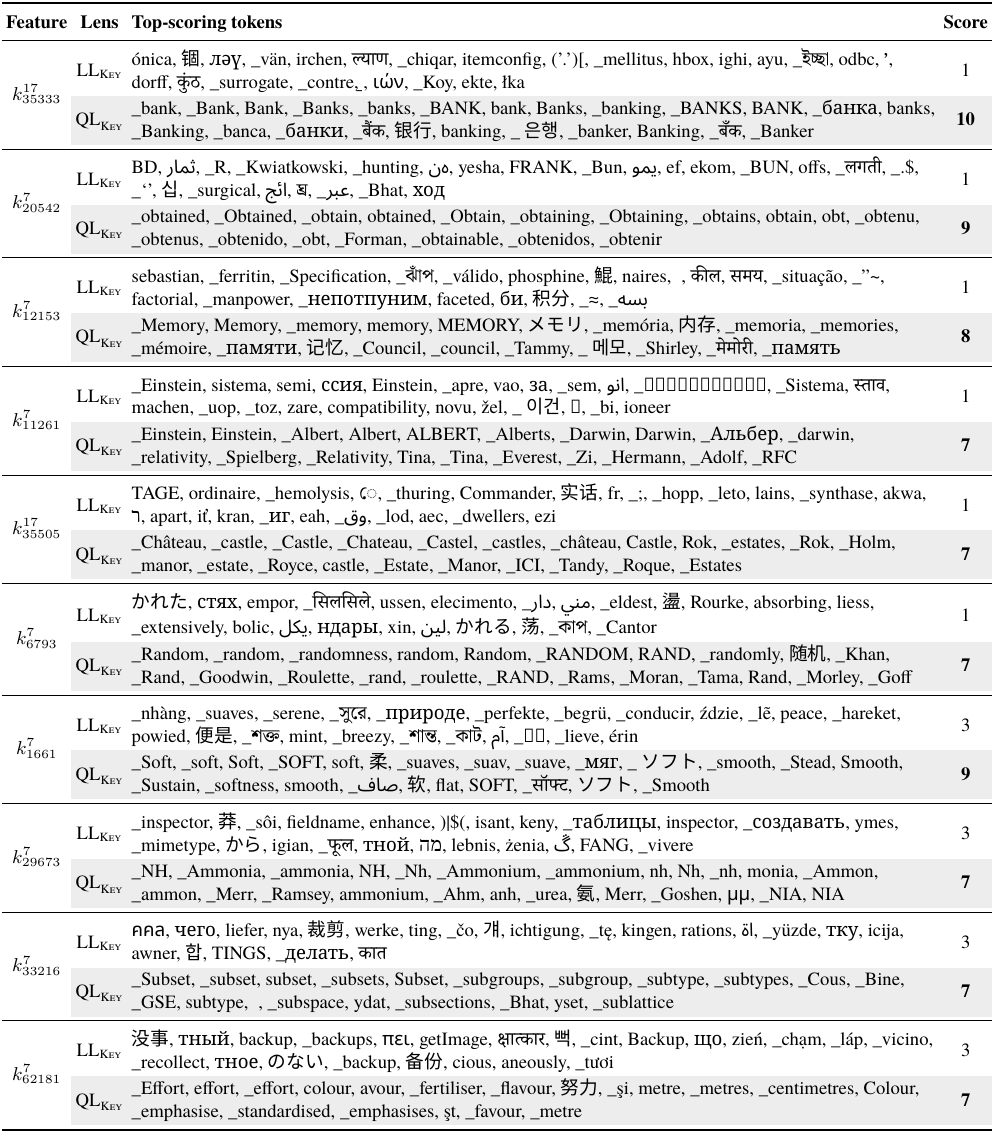}
  \caption{Qualitative \textbf{key} feature examples on Gemma-3-1B (65K).}
  \label{fig:qualJ-g1b-key}
\end{figure*}

\begin{figure*}[p]
  \centering
  \includegraphics[width=\linewidth]{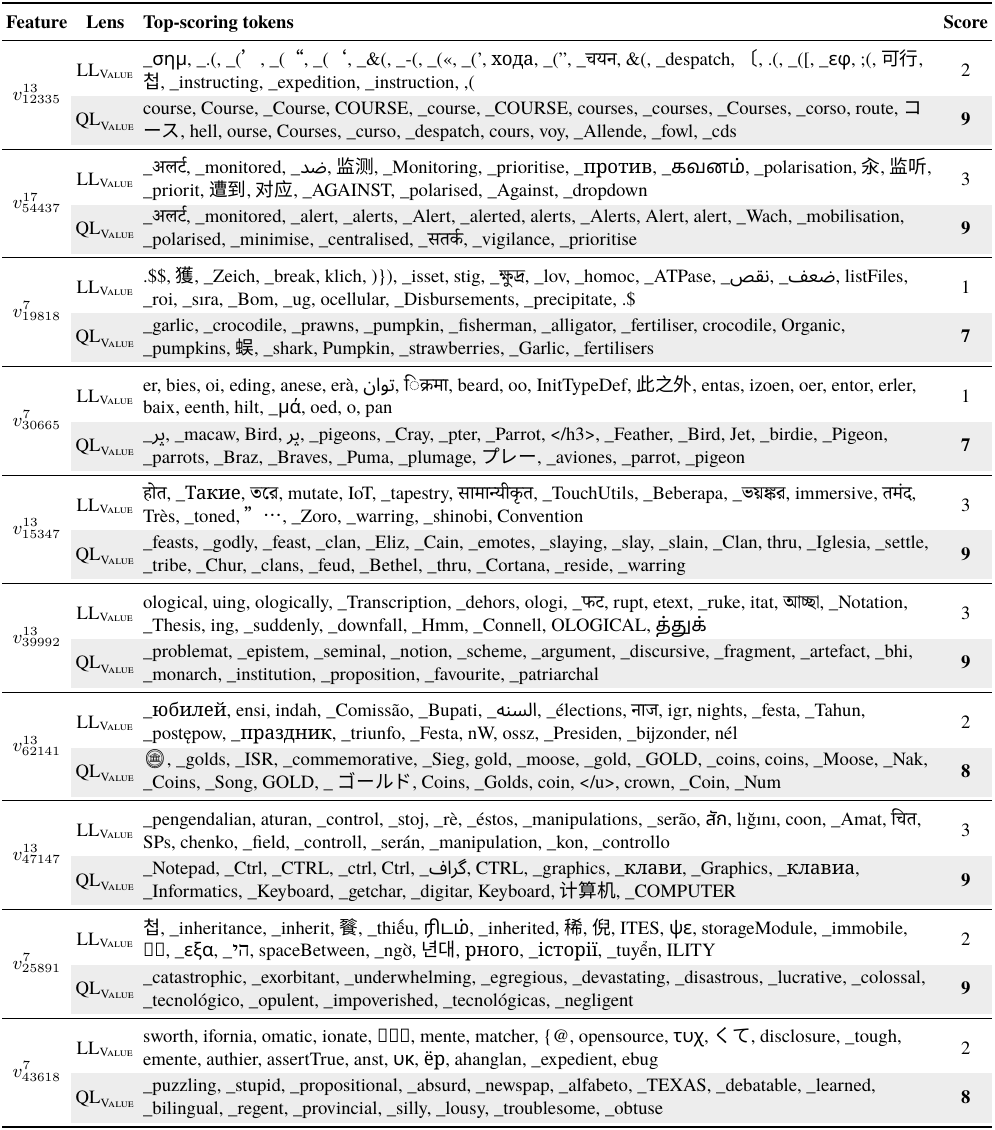}
  \caption{Qualitative \textbf{value} feature examples on Gemma-3-1B (65K).}
  \label{fig:qualJ-g1b-value}
\end{figure*}

\begin{figure*}[p]
  \centering
  \includegraphics[width=\linewidth]{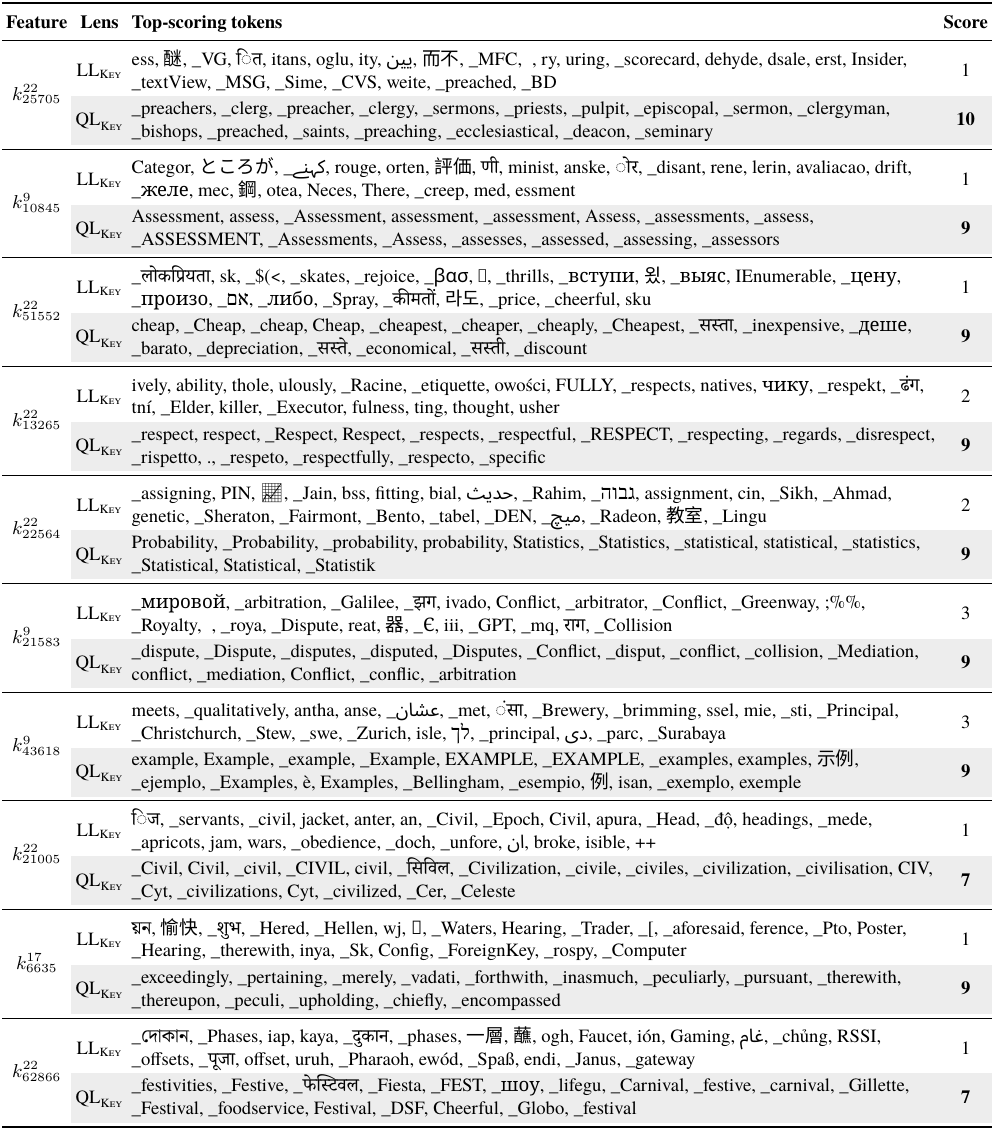}
  \caption{Qualitative \textbf{key} feature examples on Gemma-3-4B (65K).}
  \label{fig:qualJ-g4b-key}
\end{figure*}

\begin{figure*}[p]
  \centering
  \includegraphics[width=\linewidth]{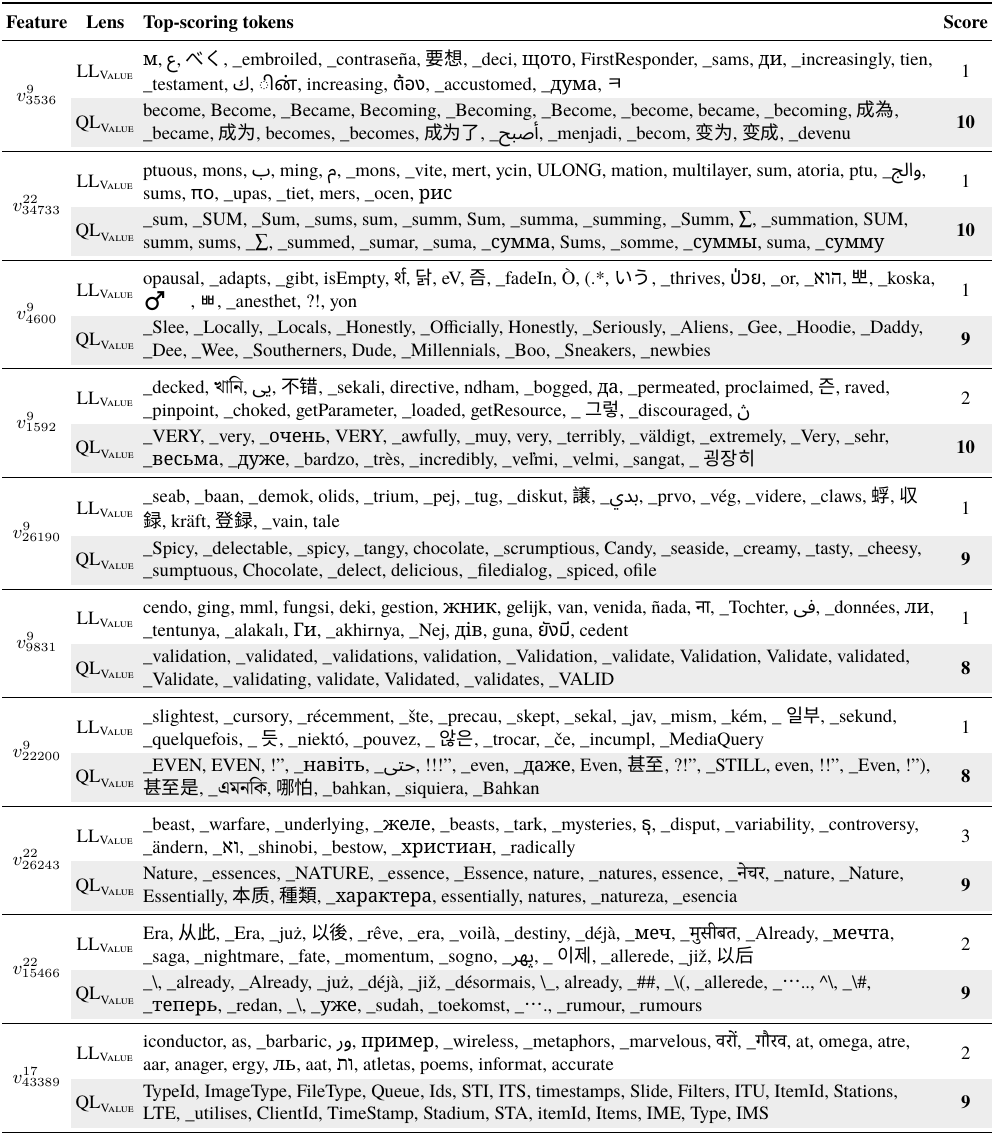}
  \caption{Qualitative \textbf{value} feature examples on Gemma-3-4B (65K).}
  \label{fig:qualJ-g4b-value}
\end{figure*}

\begin{figure*}[p]
  \centering
  \includegraphics[width=\linewidth]{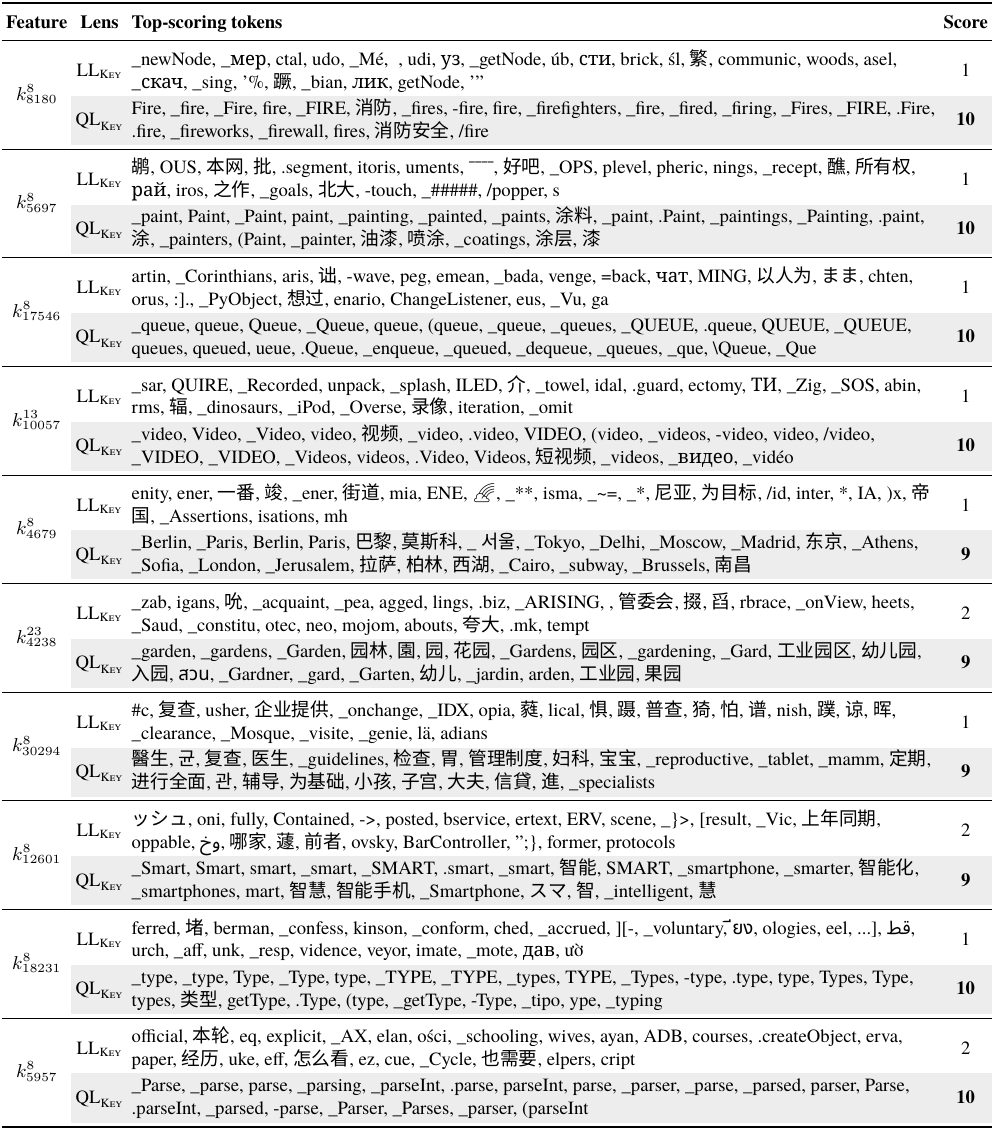}
  \caption{Qualitative \textbf{key} feature examples on Qwen-3-1.7B-Base (32K).}
  \label{fig:qualJ-qwen-key}
\end{figure*}

\begin{figure*}[p]
  \centering
  \includegraphics[width=\linewidth]{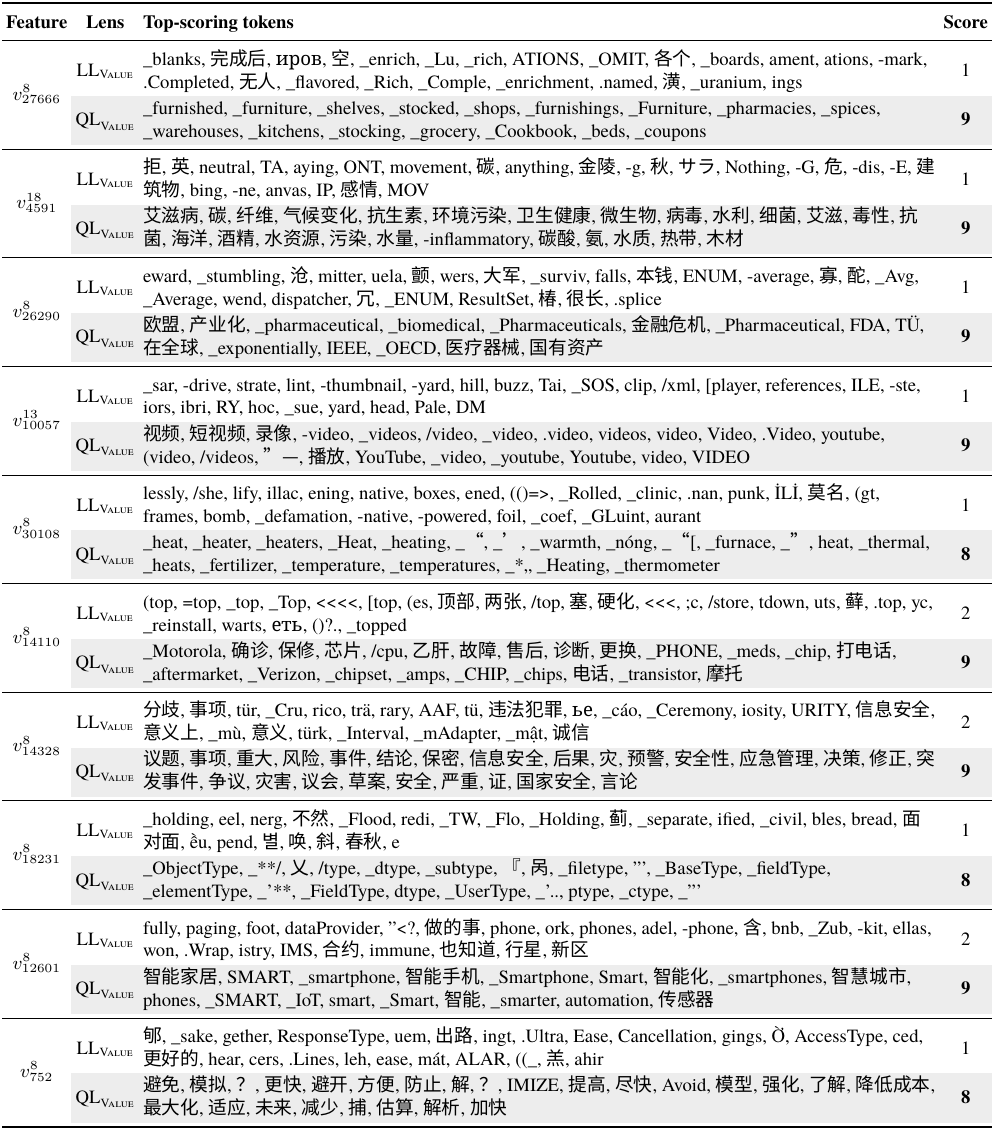}
  \caption{Qualitative \textbf{value} feature examples on Qwen-3-1.7B-Base (32K).}
  \label{fig:qualJ-qwen-value}
\end{figure*}

\begin{figure*}[p]
  \centering
  \includegraphics[width=\linewidth]{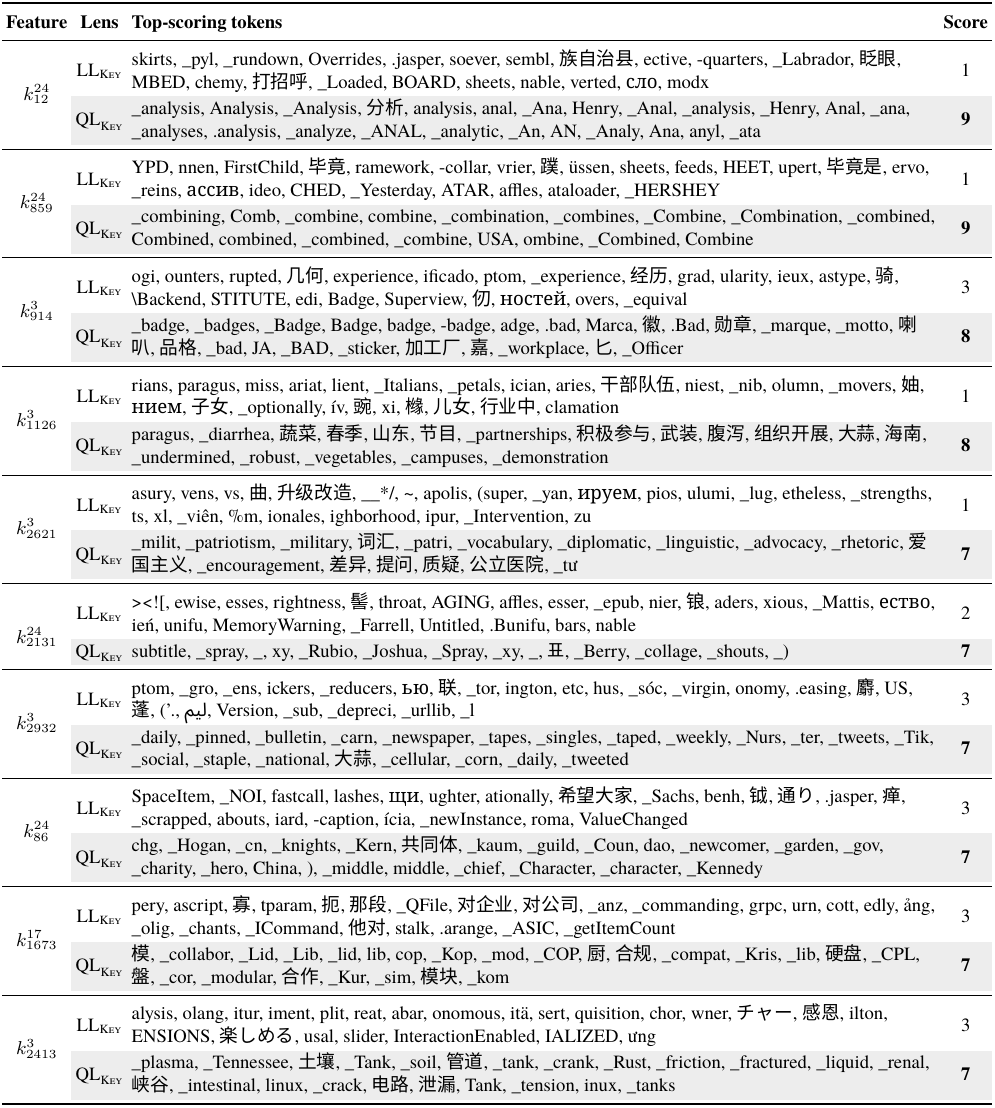}
  \caption{Qualitative \textbf{key} feature examples on Qwen-3-0.6B (transcoder).}
  \label{fig:qualJ-qwen06b-key}
\end{figure*}

\begin{figure*}[p]
  \centering
  \includegraphics[width=\linewidth]{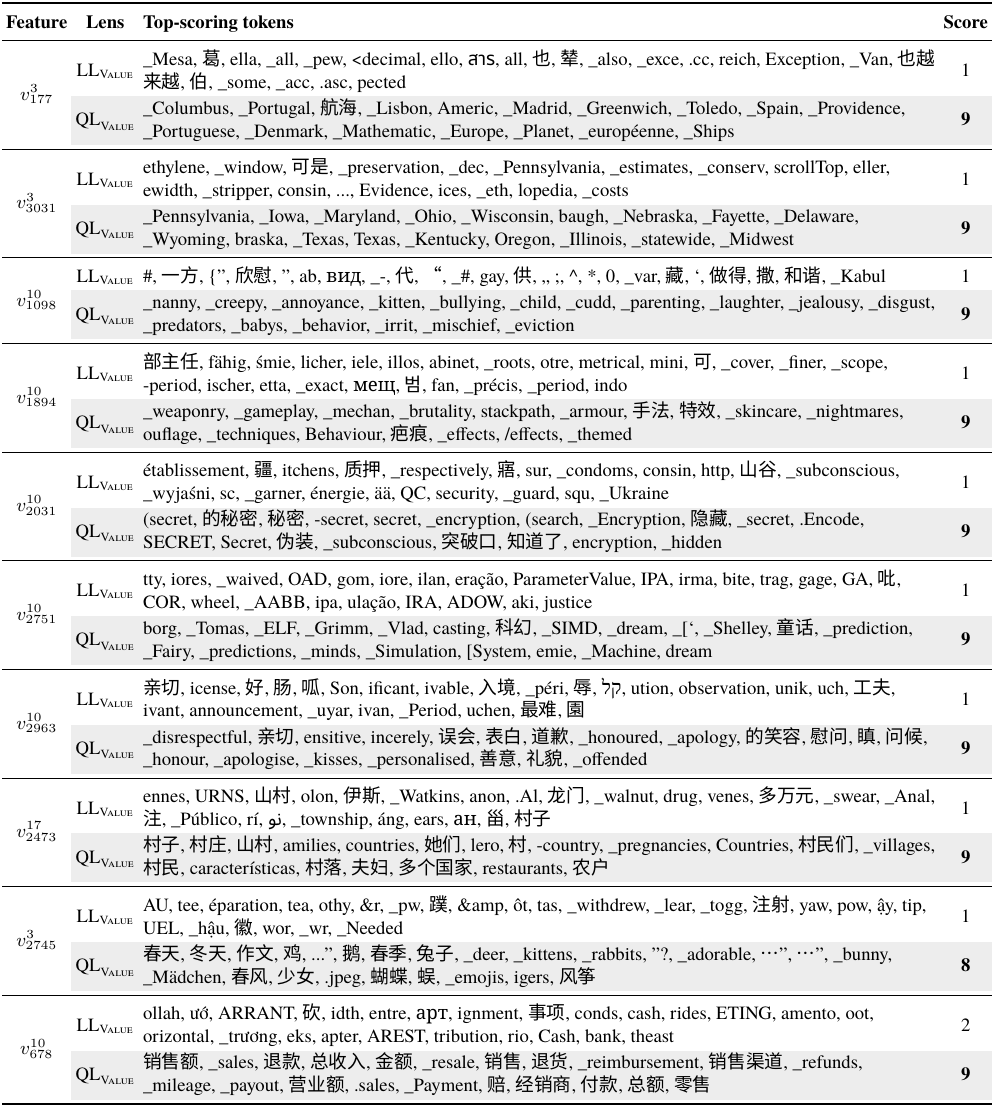}
  \caption{Qualitative \textbf{value} feature examples on Qwen-3-0.6B (transcoder).}
  \label{fig:qualJ-qwen06b-value}
\end{figure*}

\begin{figure*}[p]
  \centering
  \includegraphics[width=\linewidth]{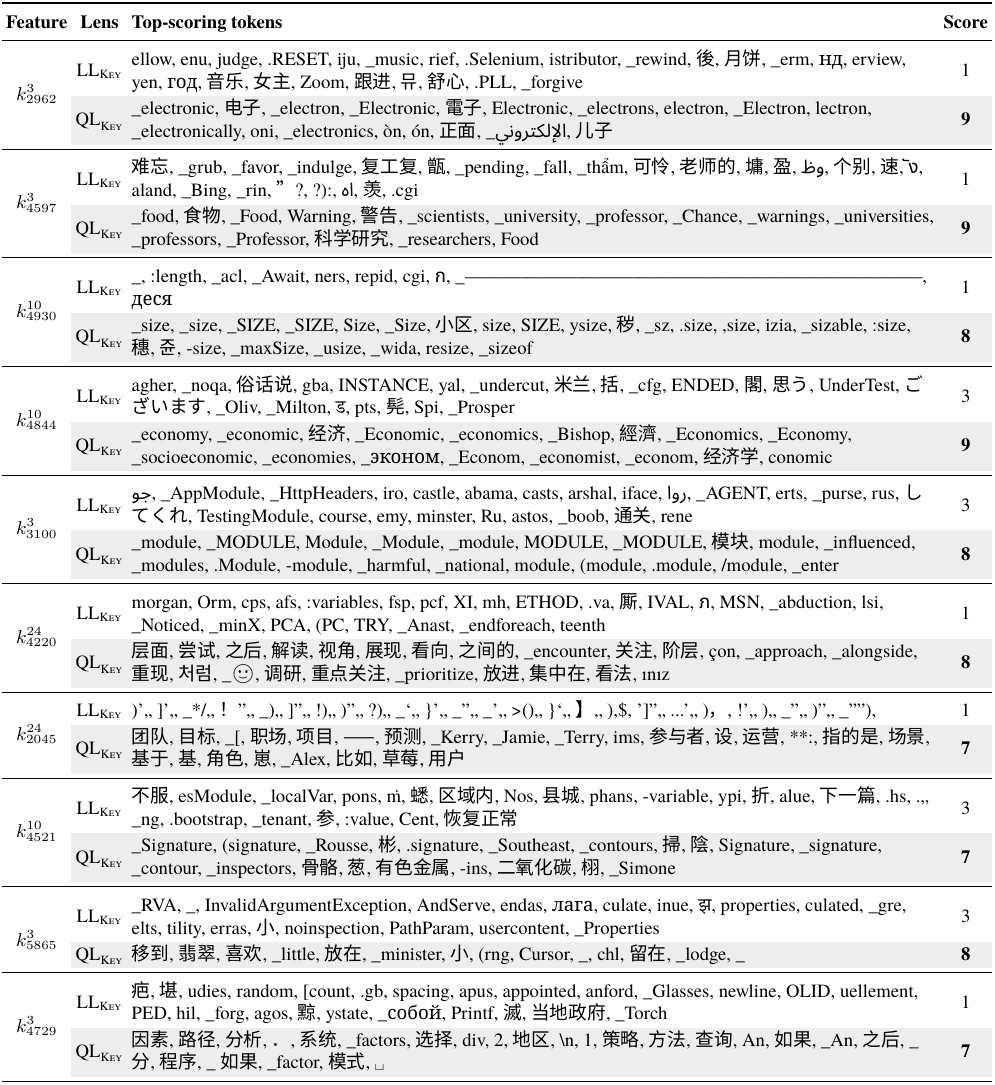}
  \caption{Qualitative \textbf{key} feature examples on Qwen-3-1.7B (transcoder).}
  \label{fig:qualJ-qwen17btc-key}
\end{figure*}

\begin{figure*}[p]
  \centering
  \includegraphics[width=\linewidth]{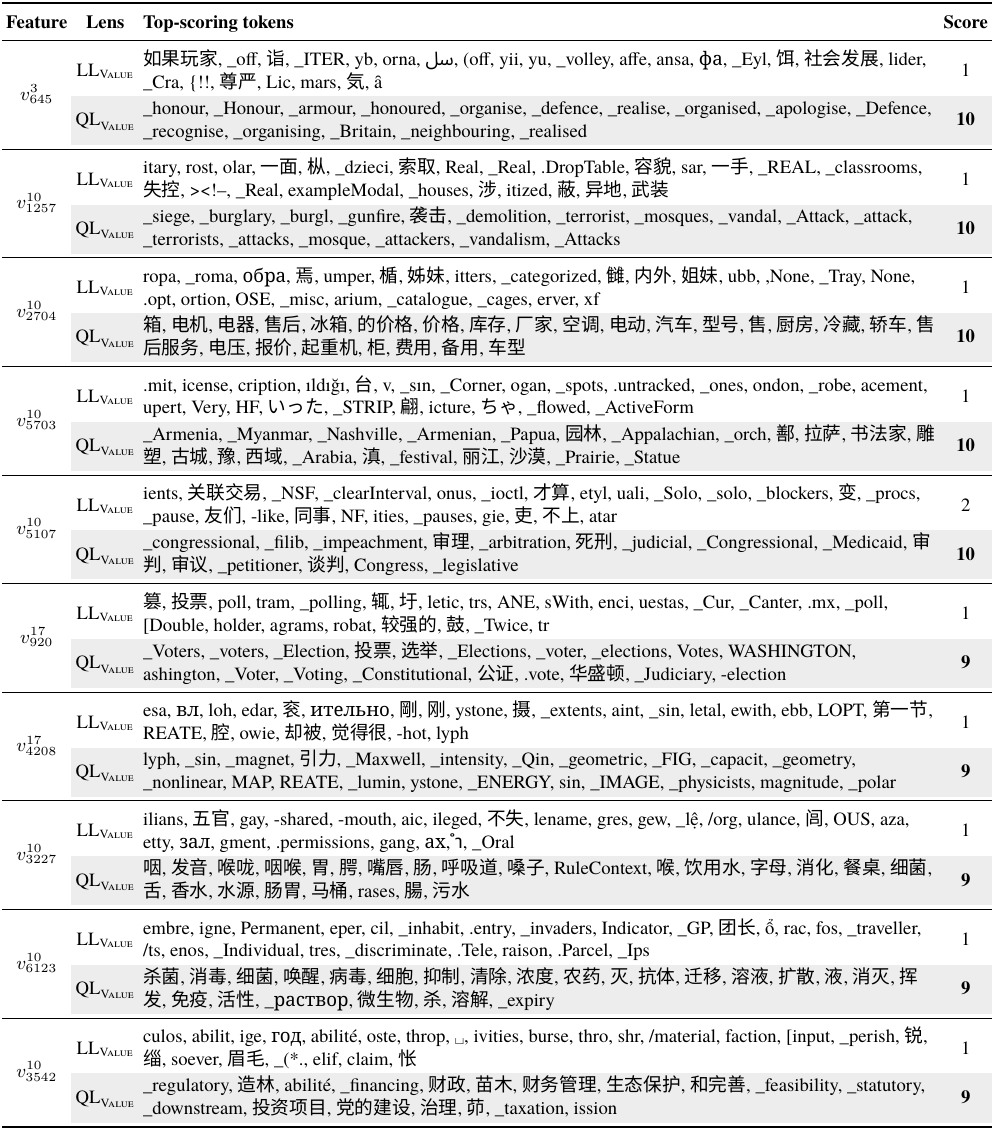}
  \caption{Qualitative \textbf{value} feature examples on Qwen-3-1.7B (transcoder).}
  \label{fig:qualJ-qwen17btc-value}
\end{figure*}

\begin{figure*}[p]
  \centering
  \includegraphics[width=\linewidth]{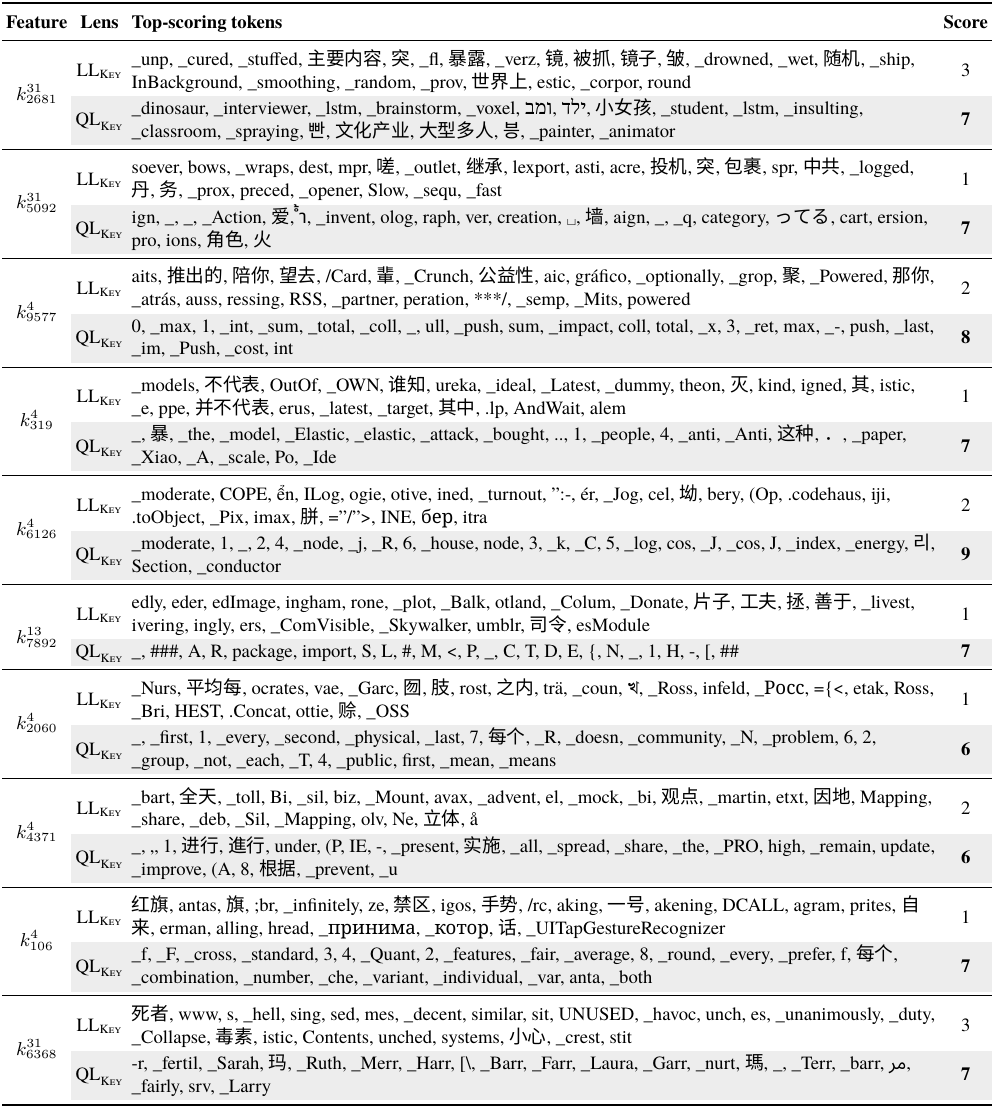}
  \caption{Qualitative \textbf{key} feature examples on Qwen-3-4B (transcoder).}
  \label{fig:qualJ-qwen4btc-key}
\end{figure*}

\begin{figure*}[p]
  \centering
  \includegraphics[width=\linewidth]{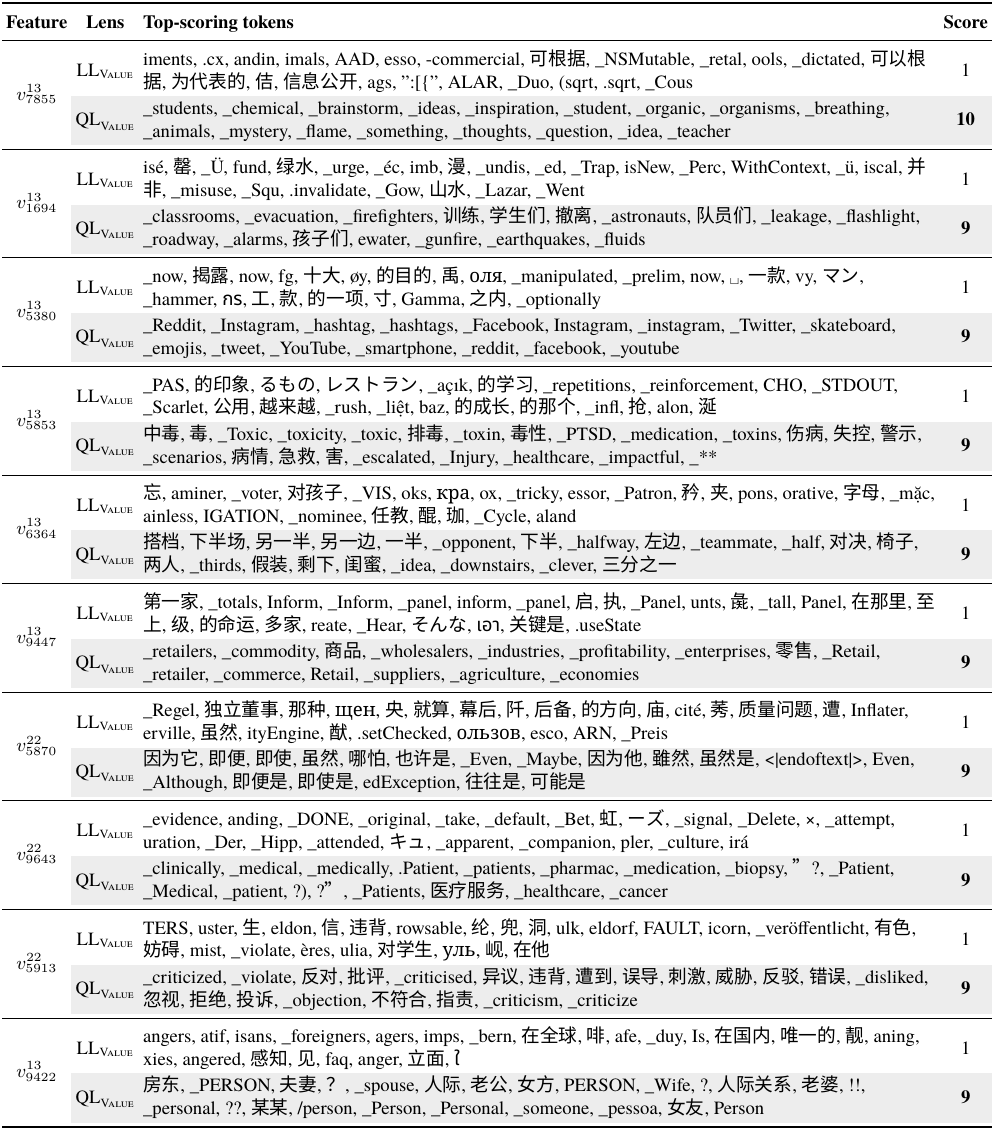}
  \caption{Qualitative \textbf{value} feature examples on Qwen-3-4B (transcoder).}
  \label{fig:qualJ-qwen4btc-value}
\end{figure*}

Section~\ref{sec:qual-eval} and Figure~\ref{tab:qual-semantic} illustrated how Query Lens token sets converge to a single recognizable concept where Logit Lens returns tokenizer fragments. Figures~\ref{fig:qualJ-gpt2-key}--\ref{fig:qualJ-qwen4btc-value} extend that comparison with more examples, spanning all eight model configurations (SAEs and transcoders) with up to ten \emph{key} and ten \emph{value} features each. Following the same format, each table lists both methods' top-scoring tokens and their GPT-5-nano interpretability scores, with the Query Lens row shaded and its score in \textbf{bold}.

\end{document}